# Graphomotor and Handwriting Disabilities Rating Scale (GHDRS): towards complex and objective assessment

Jiri Mekyska, Katarina Safarova, Tomas Urbanek, Jirina Bednarova[b], Vojtech Zvoncak, Jana Marie Havigerova, Lukas Cunek, Zoltan Galaz, Jan Mucha[a], Christine Klauszova, Marcos Faundez-Zanuy, Miguel A. Ferrer and Moises Diaz

**ABSTRACT**

Graphomotor and handwriting disabilities (GD and HD, respectively) could significantly reduce children's quality of life. Effective remediation depends on proper diagnosis; however, current approaches to diagnosis and assessment of GD and HD have several limitations and knowledge gaps, e.g. they are subjective, they do not facilitate identification of specific manifestations, etc. The aim of this work is to introduce a new scale (GHDRS – Graphomotor and Handwriting Disabilities Rating Scale) that will enable experts to perform objective and complex computer-aided diagnosis and assessment of GD and HD. The scale supports quantification of 17 manifestations associated with the process/product of drawing/ handwriting. The whole methodology of GHDRS design is made maximally transparent so that it could be adapted for other languages.

**Introduction**

The article presents the outcome of a three-year research project. The project aimed to develop a new scale for assessing graphomotor and handwriting disabilities based on objective behavioural data recorded by special software and a digitizer. In the Introduction section, different sources were reviewed to define graphomotor disabilities, and general symptoms were described to create a theoretical base. In the Methodology section, the design of the scale was explained, starting from the identification and modelling of manifestations to the calculation of normative values. Moreover, the global components of the final scale are presented. The Results section provides information about the final Graphomotor and Handwriting Disabilities Rating Scale (GHDRS) and its psychometric properties. The last part, the Discussion section, puts GHDRS in the context of previous studies and describes its limitations. As soon as a child can hold a pencil, s/he tries to reflect the real world via the graphic form. Graphomotor abilities are a set of specific psycho-motor skills comprising drawing, handwriting and writing. Creating or copying a picture is known as a process of drawing. Handwriting is the process of forming letters, symbols or figures, whereas writing comprises the composition and context of the handwritten material (Ziviani & Wallen, 2006). Current well-known theories and models of writing (Berninger & Amtmann, 2003; Cornhill & Case-Smith, 1996; Feder & Majnemer, 2007; Flower & Hayes, 1981; Kandel, Peereman, Grosjacques, & Fayol, 2011; McCloskey & Rapp, 2017; van Galen, 1991) discern two levels of the writing process: 1) higher cognitive levels, and 2) lower motor-perception levels. Thus, handwriting could be understood as the lower motor part of the complex process of writing.

It is essential for a child to successfully develop graphomotor abilities in the first years of schooling since they are necessary for later education. The important developmental milestone is handwriting automaticity. It occurs between 8–10 years when the quality and speed of handwriting increase (McCloskey & Rapp, 2017; Medwell, Strand, & Wray, 2009) and thus becomes an efficient tool for facilitating and developing ideas (Feder & Majnemer, 2007). Without automaticity, it would be difficult for the child to engage in higher-level aspects of writing. Writing disabilities are generally defined as an inability to write despite having cogni- tive potential and sufficient learning opportunities without neurological problems (Blöte & Hamstra-Bletz, 1991; Chung, Patel, & Nizami, 2020; O'Donnell & Colvin, 2019). Conventional thresholds in assessing these issues are either two standard deviations from the average achievement, or achievement two years below expected handwriting skills given the individual's chronological age (O'Hare & Brown, 1989).

Diagnostic criteria for graphomotor disabilities (GD) do not exist. Current diagnostic systems (DSM-V (American Psychiatric Association, 2013) and ICD-10 (World Health Organization, 2016)) are oriented to writing instead of handwriting issues, and they do not represent well the problems leading up to the automation of handwriting. Notably, these definitions neglect handwriting as a perceptual-motor component of the writing skill. For example, among the warning signs for writing disabilities (developmental dysgraphia) in preschool and school-aged children reported by Chung et al. (Chung, Patel, & Nizami, 2020). A majority of symptoms relate to handwriting rather than writing issues. For teenagers and adults on the other hand, the difficulties relate to ideation, syntax, grammar, and organization of thoughts. Thus, we will use the term handwriting disabilities (HD) to focus primarily on motor processes.

The prevalence of writing disabilities ranges from 7% to 34% according to country, evaluation method, and type of raters (e.g. psychologists, occupational therapists, teachers, etc.) (Cermak & Bissell, 2014; Döhla & Heim, 2016; Katusic, Colligan, Weaver, & Barbaresi, 2009). We do not have official data for handwriting issues in the Czech Republic, but a previous study showed 28.87% occurrence (Šafárová et al., 2020). There is also an agreement that boys are two to three times more frequently diagnosed with HD than girls (Katusic, Colligan, Weaver, & Barbaresi, 2009; Snowling, 2005) due to worse quality of handwriting (Hawke, Olson, Willcut, Wadsworth, & DeFries, 2009; Šafárová et al., 2020).

The research literature (Rosenblum, Weiss, & Parush, 2003) distinguishes between two major groups of symptoms used for defining and assessing poor handwriting: 1) legibility, which is usually evaluated by global or analytic scales (Feder & Majnemer, 2003; Roston Hinojosa, & Kaplan, 2008a; Šafárová, Mekyska, & Zvončák, 2021), and 2) speed understood as a number of letters per duration of handwriting/writing. Speed and legibility are not necessarily related, but there is a certain trade-off relationship (Karlsdottir & Stefansson, 2002; Weintraub & Graham, 1998).

Experts in practice use different scales or questionnaires to evaluate the final and static handwritten product. Nevertheless, these scales have been criticised because of their poor psychometric properties. Assessment based on these scales is time-consuming, they focus on only one aspect of handwriting (e.g. speed), and are not language-independent (Feder & Majnemer, 2007; Roston, Hinojosa, & Kaplan, 2008b; Šafárová, Mekyska, & Zvončák, 2021). In contrast, within the last 40 years, there has been a new approach represented by computer technology. Software and digitizers are used to record and quantify the process of handwriting itself (Faundez-Zanuy, Mekyska, & Impedovo, 2021; Šafárová, Mekyska, & Zvončák, 2021).

This approach allows researchers to assess characteristics more accurately (e.g. velocity or pressure) or to capture new ones (e.g. in-air movements, pen lifts, a tilt of the pen, etc.). Previous studies in the field of graphonomics (i.e. process approach) typically employed machine learning strategies to predict handwriting issues utilizing dynamic features as predictors (Drotár & Dobeš, 2020; Mekyska et al., 2017). Among these studies, Asselborn's research (Asselborn, Chapatte, & Dillenbourg, 2020) is groundbreaking in three aspects. They shifted the focus from binary diagnosis (children with developmental dysgraphia vs. intact children) to assessing the severity of handwriting difficulties while considering age and gender. They reduced the features of online writing to three dimensions that are independent of the product score measured by the BHK scale (Hamstra- Bletz et al., 1987). Lastly, a global score composed of four specific scores (kinematics, pressure, tilt, and static BHK product score) was created. This final step enables a more comprehensive description of the various areas of handwriting problems and facilitates focusing follow-up care accordingly.

Because HD lacks an operational definition in DSM-V and ICD-10, we defined the concept via an inventory of general symptoms. Specifications of the general symptoms were based on 1) an analysis of interviews with remedial teachers from the Czech Republic, 2) a content analysis of items from different scales concerning HD (see Appendix A1), 3) a content analysis of foreign and Czech web pages focusing on GD/HD (see Appendix A2), and 4) a literature review in the field of GD/HD research (citations are provided in the text). Figure 1 shows the process of establishing a list of general symptoms of GD/HD.

Table 1 summarises the resulting general symptoms followed by a list of their definitions. Categories "Legibility" and "Other symptoms" were not added to the following list of symptoms. We understand "Poor legibility" as a broad umbrella concept caused by distortion of handwriting, ambiguous letterforms or uneven spacing.

In the following list, each general symptom is described by symptomatic manifestation in GD/HD.

- Problems with size control. Children with GD and HD lack the consistency of letter height in their written text (Graham, Harris, & Fink, 2000; Overvelde & Hulstijn, 2011; Rosenblum, Chevion, & Weiss, 2006; Rosenblum & Dror, 2017; Rosenblum, Dvorkin, & Weiss, 2006; Rosenblum, Weiss, & Parush, 2004; Simner & Eidlitz, 2000; Smits-Engelsman & Van Galen, 1997; Wann & Jones, 1986; Wann & Kardirkmanathan, 1991) with a larger proportion of oversized letters (Rosenblum, 2008). Rosenblum and her colleagues (Rosenblum, Dvorkin, & Weiss, 2006) hypothesize that oversized letters could be an effort to create more legible text because larger letters do not require higher precision. However, Rosenblum et al. (Rosenblum & Dror, 2017) also associated overly small letters with children with HD.

*Irregular slant*. Handwritten letters of children with GD and HD do not have even inclination (Asselborn et al., 2018; Rosenblum, Chevion, & Weiss, 2006; Rosenblum, Dvorkin, & Weiss, 2006; Rosenblum, Weiss, & Parush, 2004; Smits-Engelsman & Van Galen, 1997). It should be mentioned that different writing systems have a different slant.

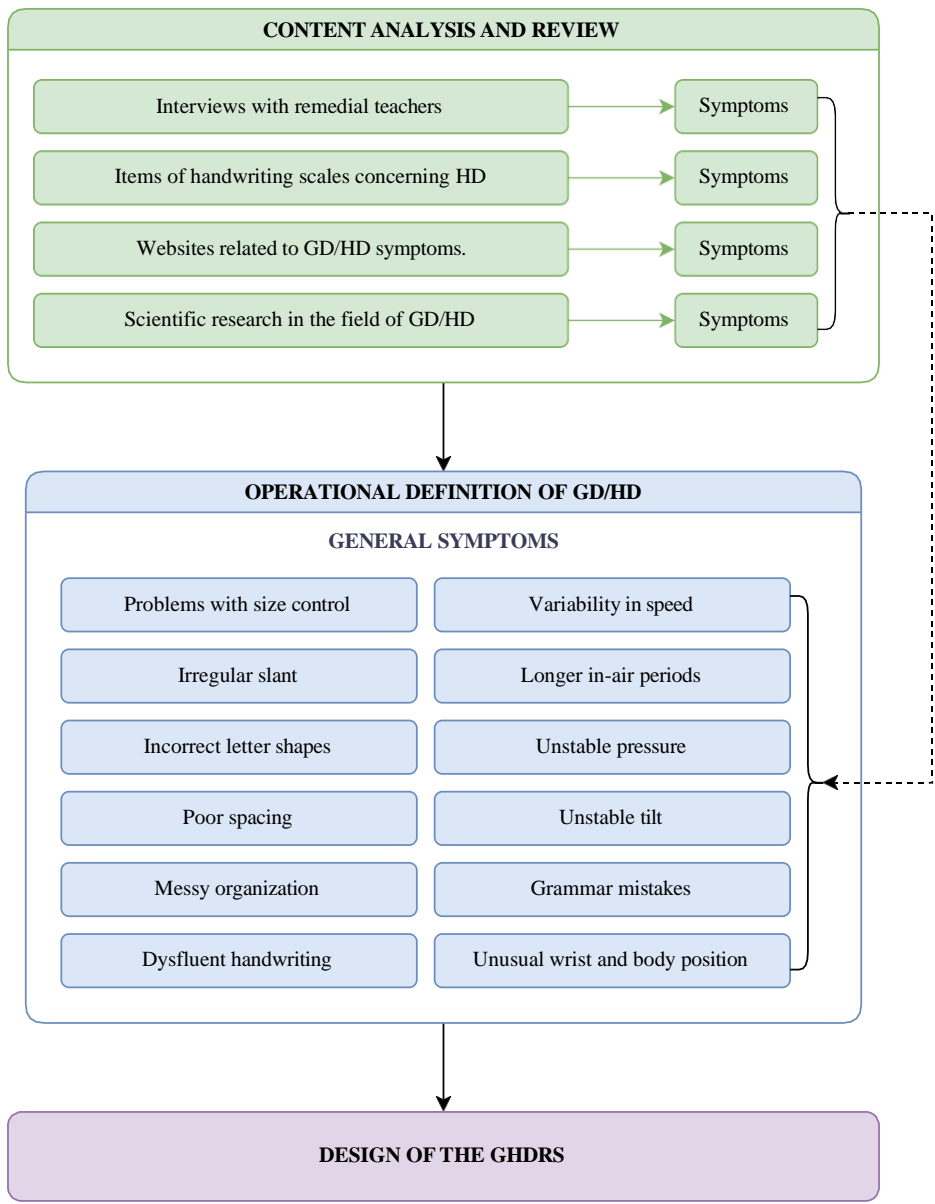

*Figure 1.* Establishing a list of general symptoms

*Table 1.* General symptoms identified in the content analysis.

| ID | General symptoms | Web pages (Appendix A1) | Scales (Appendix A2) |
|----|------------------|-------------------------|----------------------|
| 1 | Problems with size control | Variable letter size | Inconsistency in letter size |
| 2 | Irregular slant | Irregular slant | Irregular letter slant |
| 3 | Incorrect letter shapes | Incorrect letter forms | Incorrect letter form |
| 4 | Poor spacing | Incorrect spacing | Poor spacing between letters and words |
| 5 | Messy organization | Planning deficits / Poor alignment | Poor spatial organization |
| 6 | Dysfluent handwriting | Disrupted line quality / Motor deficits | Disrupted quality of movement |
| 7 | Variability in speed | Variability in speed | Slow speed |
| 8 | Longer in-air time periods | Variability in speed / Cognitive deficits | Slow speed |
| 9 | Unstable pressure | Variability in pressure | Ergonomic factors |
| 10 | Unstable tilt | Ergonomic factors | |
| 11 | Grammar mistakes | Spelling mistakes / Erasures | Grammar mistakes |
| 12 | Unusual wrist and body position | Ergonomic factors / Expression factors | Ergonomic factors |
| x | | Poor legibility | Lower legibility |
| x | | Other symptoms | |

-*Incorrect letter shapes*. Children with GD and HD have problems forming correct letter shapes (Rosenblum, Weiss, & Parush, 2004; Simner & Eidlitz, 2000; Smits-Engelsman & Van Galen, 1997), and the curvature is more pointed (Rosenblum, Dvorkin, & Weiss, 2006; Rosenblum, Weiss, & Parush, 2004; Simner & Eidlitz, 2000; Smits-Engelsman & Van Galen, 1997).

-*Poor spacing*. Children with GD and HD have bad gap layouts (Rosenblum, Chevion, & Weiss, 2006; Rosenblum, Dvorkin, & Weiss, 2006; Rosenblum, Weiss, & Parush, 2004; Smits-Engelsman & Van Galen, 1997) with over-large, missing or irregular gaps (Asselborn et al., 2018; Rosenblum, Chevion, & Weiss, 2006; Simner & Eidlitz, 2000). Consequently, the words are hardly recognised, which decreases the legibility of the whole text.

-*Messy organization*. Children with HD cannot keep text on the line (Jucovičová, 2014; Rosenblum & Dror, 2017; Rosenblum, Weiss, & Parush, 2004) and cannot stay within the margins (Engel-Yeger, Nagauker-Yanuv, & Rosenblum, 2009; Rosenblum, 2018; Rosenblum, Aloni, & Josman, 2010). Previous studies are not consistent in their outcomes and consider problems with size control, poor spacing and maintaining on the line either due to motor deficit or cognitive deficit (deficit of executive functions) (Rosenblum, 2018; Rosenblum, Aloni, & Josman, 2010; Rosenblum, Weiss, & Parush, 2003). Even though, the organisation of handwriting on the page along with velocity and correction were observed to be the most critical symptoms and manifestations distinctive of children with dysgraphia (Engel-Yeger, Nagauker-Yanuv, & Rosenblum, 2009).

-*Dysfluent handwriting*. It is usually related to the speed and automation of hand-writing, which is often not fully achieved in HD (Nicolson & Fawcett, 2011; Tucha, Tucha, & Lange, 2008). Children with GD and HD usually have shaky, jerky, rugged and rambling lines. According to Van Galen et al. (Van Galen, Portier, Smits

Engelsman, & Schomaker, 1993), children with poor handwriting show immaturity when it comes to precise and consistent movement control, called neuromotor noise. It has also been observed that children with HD write less fluently because the text divides into smaller segments (Rosenblum, Chevion, & Weiss, 2006) with greater tremor (Asselborn et al., 2018). According to Meulenbroek et al. (Meulenbroek & Van Galen, 1986) children use pen elevation and pauses for optimal speed planning, slant, following letter shape, etc. Interruption of the handwriting is typical for children with GD/HD (Chang & Yu, 2013), with more pen elevations above the surface (Rosenblum, Chevion, & Weiss, 2006). However, Paz-Villagran et al. (Paz- Villagrán, Danna, & Velay, 2014) found no differences in pen elevations between children with HD and children with typical handwriting development.

- *Variability in speed.* There is a consensus among experts that slow speed/velocity of drawing/handwriting can be observed among children with GD/HD (Jucovičová, 2014; Overvelde & Hulstijn, 2011; Prunty & Barnett, 2017; Rosenblum, 2018; Rosenblum, Chevion, & Weiss, 2006; Rosenblum, Weiss, & Parush, 2003; Tseng & Chow, 2000). However, research outcomes in this manifestation are less conclusive. A few studies even found that children with non-proficient handwriting are faster than children without impairment in easy graphomotor tasks (e.g. four oblique lines (Razian, Fairhurst, & Hoque, 2004); or flower-trail drawing (Smits- Engelsman, Niemeijer, & van Galen, 2001)). A great number of studies agree on the overall slower performance of children with GD/HD (Biotteau et al., 2019; Feder & Majnemer, 2007; Graham, Struck, Santoro, & Berninger, 2006; Hamstra-Bletz et al., 1987; Kaiser, Albaret, & Doudin, 2009; Overvelde & Hulstijn, 2011; Parush, Lifshitz, Yochman, & Weintraub, 2010; Rosenblum, Parush, & Weiss, 2003; Schoemaker & Smits- Engelsman, 1997; Smits-Engelsman, Niemeijer, & van Galen, 2001; Søvik, Arntzen, & Thygesen, 1987). Moreover, children with HD have lower average stroke velocity (Khalid, Yunus, & Adnan, 2010) and make acute turns in velocity within individual strokes (Asselborn et al., 2018). The lower velocity of drawing/handwriting strongly correlates with the longer time needed to accomplish the task, especially for creating more difficult letters (Rosenblum, Chevion, & Weiss, 2006; Rosenblum & Roman, 2009). Engel-Yeger et al. (Engel-Yeger, Nagauker-Yanuv, & Rosenblum, 2009) attri- bute the longer time to excessive correction, erasing, or deletion, which they believe results from poor planning and poor ability to distinguish between details (whether visual or phonetic). On the other hand, other studies do not find any differences between children with and without GD/HD (Engel-Yeger, Nagauker-Yanuv, & Rosenblum, 2009; Hamstra-Bletz & Blöte, 1993; Khalid, Yunus, & Adnan, 2010; Rubin & Henderson, 1982; Schoemaker, Schellekens, Kalverboer, & Kooistra, 1994; Søvik, Arntzen, & Thygesen, 1987; Søvik, Flem Mæland, & Karlsdottir, 1989; Wann & Jones, 1986). It seems that variability in speed during handwriting is a better indi- cator of GD/HD (Rubin & Henderson, 1982; Wann & Jones, 1986). This is consistent with the outcomes of Kushki et al. (Kushki, Schwellnus, Ilyas, & Chau, 2011), who claimed that children who write slower do not have to be poor handwriters. Nevertheless, with time constraints during exams or assignments, a slower tempo may affect a child's academic performance. Further, studies do not agree on whether girls are faster writers (Graham, Weintraub, & Berninger, 1998; Ziviani, 1984) than boys (van Galen, 1991) or whether there is no difference (Wicki, Lichtsteiner, Geiger, & Müller, 2014).

*- Longer in-air time periods*. Digitizers allow recording movement of the pen above the surface. Children with HD spend more time with the tip of a writing instrument in the air than children without handwriting issues (Asselborn et al., 2018; Rosenblum & Dror, 2017; Rosenblum, Weiss, & Parush, 2003; Rosenblum, Weiss, & Parush, 2004), even within a single letter (Rosenblum, Dvorkin, & Weiss, 2006; Rosenblum, Parush, & Weiss, 2003). In addition, further examination of the handwriting movements above the surface indicates that a child does not keep the pencil in the air at rest (Rosenblum, Parush, & Weiss, 2003; Rosenblum, Weiss, & Parush, 2004). Thus, the in- air trajectory of children with HD is longer, and the ratio to the overall path of pen movement during writing increases with the length of the text and the unfamiliarity of the written shapes (Rosenblum, Parush, & Weiss, 2003).

Further, more time spent in-air could be linked to difficulties recalling a correct letter shape as a result of poor orthographic coding process in working memory (Döhla & Heim, 2016; Romani, Ward, & Olson, 1999; Rosenblum & Dror, 2017). It is important to note that most research of the in-air time has been conducted for Hebrew, where the individual letters are not connected. Therefore, there is a possibility that the in-air time will not be so significant in other languages (Rosenblum, Parush, & Weiss, 2003).

*- Unstable pressure*. A common manifestation linked by remedial teachers to GD/HD is excessively high pressure on the surface. In contrast to observations by practitioners, researchers observed differences in variability of the pressure among children with GD/HD (Khalid, Yunus, & Adnan, 2010; Rosenblum & Dror, 2017). Kushki et al. (Kushki, Schwellnus, Ilyas, & Chau, 2011) find that pressure on the surface increases with the length of writing in both groups of children. Rosenblum et al. (Rosenblum, 2018) did not find a relationship between the pressure exerted on the surface among children with HD but observed a relationship between pressure and a set of processes that reflect organizational skills, working memory and emotional control.

*- Unstable tilt.* When writing, children with GD/HD start with a significantly larger angle with the surface, up to $90°$, compared to the recommended $45°$ (Rosenblum, Chevion, & Weiss, 2006). According to Asselborn et al. (Asselborn et al., 2018), children with typical handwriting development are more flexible in tilt within one stroke. Children with HD "convulsively" grip the pen at the same angle throughout the handwriting movements. In contrast to these findings, other authors observed variation in the tilt during the handwriting in children with HD (Mekyska et al., 2017; Rosenblum, Chevion, & Weiss, 2006).

*- Grammar mistakes*. The written text of children with HD does not correspond with their grade level (Kandel, Lassus-Sangosse, Grosjacques, & Perret, 2017; Rodríguez & Villarroel, 2016; Rosenblum & Dror, 2017). Usually, children with HD have problems with differentiating similar letters (Döhla & Heim, 2016; Jucovičová, 2014; Romani, Ward, & Olson, 1999) and spelling (Döhla & Heim, 2016; Romani, Ward, & Olson, 1999; Rosenblum & Dror, 2017). This symptom is characteristic of the writing process more than handwriting itself. However, with increasing mistakes, erases, crossing out and correcting the text occurs in children with HD (Engel-Yeger, Nagauker-Yanuv, & Rosenblum, 2009; Rosenblum, 2018; Rosenblum, Weiss, & Parush, 2004). According to Rosenblum et al. (Rosenblum, 2018), this symptom relates to difficulties in plan- ning abilities.

- *Unusual wrist and body position.* Some authors report a relationship between handwriting disabilities and incorrect sitting position (Rosenblum, Goldstand, & Parush, 2006; Smith-Zuzovsky & Exner, 2004). Particularly the effect of poor position on manipulating the pen in the palm and the possible effect on attention, which is disturbed by the repeated search for a more suitable stable position when writing. Overvelde et al. (Overvelde & Hulstijn, 2011) argue however that even if it has repeatedly been pointed out that there is a relationship between sitting position and HD, a causal link has not yet been demonstrated. Similarly, the influence of incorrect grip on handwriting development has not been confirmed (Burton & Dancisak, 2000; Graham & Weintraub, 1996; Schwellnus et al., 2012). Nevertheless, it is likely, that immature grip or incorrect body position will cause fatigue or pain leading to aversion to handwriting (Engel-Yeger, Nagauker-Yanuv, & Rosenblum, 2009; Pokorná, 2004; Rosenblum, Weiss, & Parush, 2003; Zelinková, 2015). Although children with GD/HD experience fatigue and pain sooner than their intact classmates (Parush, Pindak, Hahn-Markowitz, & Mazor-Karsenty, 1998), most of the research does not address this (Overvelde & Hulstijn, 2011).

GD and HD could have a detrimental impact on the child's quality of life, and without a proper diagnosis, it is hard to find appropriate remediation. Nevertheless, current approaches to diagnosis and assessment of GD and HD have several limitations and knowledge gaps. As was mentioned, commonly used questionnaires and scales have several drawbacks (Rosenblum, Weiss, & Parush, 2003) and rely on the subjective assessment of the expert.

Although research of objective parameters quantifying drawing/handwriting/GD/HD exists (in the field of graphonomics) (Kao, Hoosain, & Van Galen, 1986; Van Gemmert & Contreras-Vidal, 2015; Van Gemmert & Teulings, 2006), prior studies were usually based on one type of graphomotor element, e.g. loops or ellipses (Bosga-Stork, Bosga, & Meulenbroek, 2017; Pellizzer & Zesiger, 2009), or one word/sentence (e.g. BHK test or MHA test (Asselborn et al., 2018; Falk, Tam, Schellnus, & Chau, 2011)). Moreover, authors compared children with typical handwriting development to children with HD using statistical tests such as ANOVA or t-test (Engel-Yeger, Nagauker-Yanuv, & Rosenblum, 2009; Paz-Villagrán, Danna, & Velay, 2014; Van Galen, Portier, Smits-Engelsman, & Schomaker, 1993).

In other words, prior studies remain mainly in the comparative research design, and except for one pioneering publication (Asselborn, Chapatte, & Dillenbourg, 2020), one cannot identify any scales that would enable objective (computerized) diagnosis of GD/ HD. Finally, to the best of our knowledge, there does not exist any methodology for detailed and objective assessment of the specific manifestations associated with GD/HD. The current study aims to bridge the above-mentioned knowledge gaps, address the limitations of current approaches, and introduce a new Graphomotor and Handwriting Disabilities Rating Scale (GHDRS) that will enable experts to perform objective and complex computer-aided diagnosis and assessment of GD and HD. Although the scale has been primarily developed for the Czech language, we made all the methodology. maximally transparent so that it could also be adapted to other languages.

## 2. Methodology

This section describes the design of the GHDRS and a dataset that was used to create normative data. Nevertheless, to better understand some steps during the design we firstly illustrate a way the scale will be used during the assessment of GD/HD.

*2.1. General concept of the GHDRS-based assessment*

Assessment based on the newly designed GHDRS will be done in the following steps (description of use case), see Figure 2:

- A child will perform several drawing/handwriting tasks (the set of tasks will depend on the grade level) on paper using an inking pen.

- The paper will be overlaid on a digitizer that will record the process of handwriting/ drawing. Consequently, the process will be automatically parameterised using signal processing algorithms.

- Parameters will be automatically fed into mathematical models. Each model will be associated with one manifestation (e.g. dysfluency) and will rate its severity (in relation to normative data).

- Utilising the automatic mathematical modelling, an expert (e.g. a remedial teacher) will have access to a complex profile of the child's performance, i.e. s/he will know whether the child has some manifestations of GD/HD and how severe they are (s/ he will see the position of the child in a probability density function).

- Alongside the detailed profile, the expert will be provided with global scores of GD/ HD.

Except for the first step, everything will be done automatically. The expert will use the profile and the global scores e.g. to make a diagnosis, monitor the progress of the child, introduce targeted and effective therapy, etc.

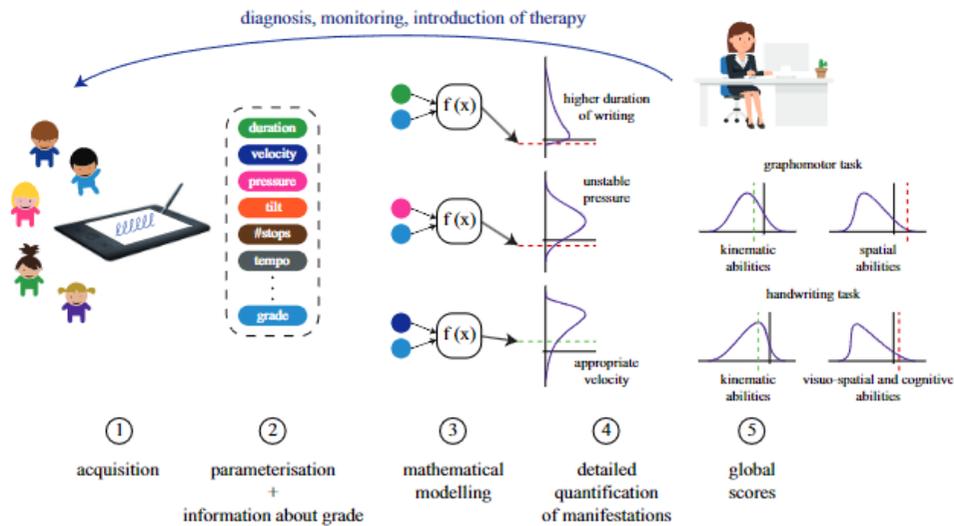

*Figure 2. Assessment based on the GHDRS.*

*2.2. Database*

The whole database (available upon request) consists of several sources of data: 1) socio- demographic data (e.g. sex, age, grade etc.), 2) scores of three questionnaires: Handwriting Proficiency Screening Questionnaires for Children (HPSQ – C) (Rosenblum & Gafni-Lachter, 2015; Šafárová et al., 2020), Handwriting legibility scale (HLS) (Barnett, Prunty, & Rosenblum, 2018), shortened version of Concise Assessment Methods of Children Handwriting (SOS: BHK) (Van Waelvelde, Hellinckx, Peersman, & Smits- Engelsman, 2012), 3) drawing/handwriting data acquired by a digitizer and 4) expert evaluation.We used a convenience sample of 353 children from the final grade of kindergarten to the fourth grade of elementary school. In the Czech Republic, children start compulsory schooling at the age of 6, so the final grade of kindergarten is usually between the ages of 5 and 6. Participants were enrolled in 8 kindergartens, 14 elementary schools, and 2 counselling centres in the Czech Republic, covering 5.95% (N = 21) of which were children with HD while 82.44% (N = 291) were children with typical handwriting development (THD; intact). Children with HD were enrolled from both, schools and counselling centres. Data about diagnosis was missing from 11.61% (N = 41) children. The average age of the whole cohort was 8 years and 18 months (SD = 19.4 months). Over half the sample (56.09%) were boys.

Detailed information on grade, sex and diagnosis of developmental dysgraphia are reported in Table 2. Parents of all children participating in this study signed an informed consent form approved by the Research Ethics Committee of the Masaryk University (approval number: EKV-2017-020-R1). Throughout the entire duration of this study, we strictly followed the Ethical Principles of Psychologists and Code of Conduct released by the American Psychological Association (https://www.apa.org/ethics/code/).

Because of subjectivity and variability in the diagnostic process and missing data about diagnosis, we asked an expert in the field to evaluate stimuli on a scale from zero to four, where 0 indicates intact child (N = 7), 1 indicates a child with subtle graphomotor issues or handwriting specificity (N = 38), 2 indicates a child with graphomotor issues or partial handwriting specificity (N = 128), 3 indicates a child with handwriting disabilities (N = 127) and 4 indicates a child with severe handwriting disabilities (N = 53). The value of McDonald's ω for evaluation of graphomotor elements is .88 (95% CI [0.86; 0.90]).

Children from the first to the fourth grade (N = 207) filled out HPSQ – C. This self-evaluation method yields the child's point of view. A study on a Czech sample (Šafárová et al., 2020) showed acceptable reliability (McDonald's ω = .7) and construct validation through confirmatory factor analysis (CFA: $\chi^2(32)$ = 31.12, p = .51; CFI = 1.0; TLI = 1.0; RMSEA = 0.0; SRMR = 0.04) supported a three-factor structure. A higher score in the questionnaire refers to more severe HD. The prior Czech study established 19 points as the cut-off for HD. The mean score of HPSQ – C in the present study was 13.4 (SD = 5.8). The mean score of children with diagnosis of DD was 18 (SD = 5.5) and the mean score for children with THD was 12.7 (SD = 5.5). There was a significant difference between those two groups (t(183) = −3.76, p < .01, d = −0.96).

The transcription task was scored using HLS and BHK scales on a sample of 103 children from third and fourth grades who performed the same handwriting task. Significant differences were found between children diagnosed with dysgraphia (MHLS = 13.9, SD = 3.6; MBHK = 14.2 SD = 3.1) and children with typical development of handwriting (MHLS = 8.9, SD = 2.9; MBHK = 10.1, SD = 3.3) were found for HLS (t(97) = 5.58, p < .001, d = 1.66) and SOS: BHK (t(97) = 4.19, p < .001, d = 1.25). Thus, children with diagnosed dysgraphia obtained higher scores in all questionnaires implying worse handwriting performance.

All children were asked to perform a specifically designed protocol consisting of 7 elementary graphomotor tasks (TSK1 – Archimedean spiral (approximately 15 cm in height); TSK2 – half-sized version of TSK1; TSK3 – upper loops; TSK4 – lower loops; TSK5 – zig-zag line; TSK6 – arcade; TSK7 – a combination of TSK3 and TSK4) and one paragraph copying task, whose content was depended on the grade of a child. Regarding the graphomotor part of the protocol, it was designed in a way so that the tasks cover the building blocks of letters used in the Latin alphabet (see Figure 4; a printable version could be found in the Appendix A3). In the Czech Republic, a child should master these elements before s/he enters the 1st grade of a primary school. All graphomotor elements were printed on paper and the children were asked to copy them.

For the handwriting part of the protocol, children were asked to copy paragraphs reported in Table 3. The paragraphs were printed on a template using block letters, but the children were asked to copy them in cursive ones. Children attending the 1st or 2nd grade were writing on a template consisting of lines printed 20 mm apart. On the other hand, children attending the 3rd/4th grade used a template where the lines were printed 15 mm apart. These line spacings are equivalent to the ones used by children at Czech schools. The templates could be found in the appendices A4 and A5.

The children were writing on a paper that was overlaid on and fixed to a digitising tablet Wacom Intuos Pro L (PHT-80). For this purpose, they used a Wacom Inking Pen that enabled them to have immediate visual feedback when writing/drawing and to replicate the feel of writing with a conventional inking pen. Data were recorded using the freely available HandAQUS acquisition software (Mucha, Mekyska, Zvoncak, Galaz, & Smekal, 2022).

*Table 3. Handwriting part of the acquisition protocol.*

| TSK8, 1st grade (original) | TSK9, 2nd grade (original) | TSK10, 3rd/4th grade (original) |
|---|---|---|
| Hana ráda maluje. | Brzy bude jaro. | Gusta, Lenka, Hana a Stáňa jsou spolužáci. |
| Banány vybarví žlutě. | Sluníčko již hřeje. | Brzy je čeká vysvědčení. |
|  | Gusta a Hana tancují. | Po prázdninách budou chodit do čtvrté třídy. |
| **TSK8, 1st grade (translation)** | **TSK9, 2nd grade (translation)** | **TSK10, 3rd/4th grade (translation)** |
| Hana likes to paint. | Spring is coming soon. | Gusta, Lenka, Hana and Stáňa are classmates. |
| She will paint bananas in yellow. | The sun is already warming. | They will soon have a report card. |
|  | Gusta and Hana are dancing. | After the holidays, they will attend the fourth grade. |

*2.3. Handwriting disabilities criterion*

We repeatedly encountered a substantial issue regarding external validation criteria during the whole study. There was reasonable doubt surrounding the reliability of diagnoses because of the lack of diagnostic criteria for DD. Discrepancies in the evaluation of children with DD in different districts of the Czech Republic were confirmed by remedial teachers. In addition, it seemed that children's self-evaluation should also be considered alongside other data. Therefore, a new composite score was created to indicate GD/HD in the sample.

The new handwriting disabilities criterion (HDC) combines overall expert evaluation (OEE) and with the child's self-evaluation (HPSQ – C). A one-way analysis of variance (ANOVA) demonstrated that the effect of OEE was significant for HPSQ – C with medium effect ($F(4, 202) = 10.39$, $p < .001$, $\omega^2 = .15$). Levene's homogeneity tests were non- significant ($p = .20$). Post-hoc testing using Holm correction did not find differences between the group of children evaluated as intact (0) and and those with subtle graphomotor issues (1), ($p = .94$), nor with children with graphomotor issues (2), ($p = .94$). Likewise, there was no significant difference between groups 1 and 2 ($p = .42$). Thus, children evaluated with 0, 1 or 2 points could therefore be regarded as one group.

Moreover, there were no significant differences between children with HD (3) and children with severe HD (4) (p = 0.94). Significant differences were found between the group of children with severe disabilities (4) on the one hand, and children with graphomotor issues (2) (p < .001), children with subtle graphomotor issues (1), (p < .001), and intact children (0) (p = .02) on the other. Similarly, significant differences were found between children with HD (3) when compared with children with graphomotor issues (2) (p < .001), children with subtle graphomotor issues (1) (p < .001), and intact children (0) (p = 0.048). Higher scores on OEE corresponded with higher mean scores of HPSQ – C.

Detailed post-hoc testing of ANOVA served as a guide for creating the final HDC. Categorical levels were identified based on thresholds for of OEE and HPSQ – C (see OEE(t) and HPSQ – C(t) in Table 4). When a child was scored 0 to 2 points by the assessment expert, we considered this to indicate a child with THD (OEE(t) = 0). A score of 3 or 4 points on the scale was taken as an indication of GD/HD (OEE(t) = 1). When a child reaches the cut-off score of 19 points in HPSQ – C, we interpreted this as an indication that the child had GD/HD (HPSQ – C(t) = 1). The scores along with the corresponding final HDC is reported in Table 4.

### 2.4. Design of the GHDRS

For better orientation, the overall design of the GHDRS is summarized in Figure 3. The blocks in the diagram contain numbered sections corresponding to the text below.12J. MEKYSKA ET AL.

*Table 4. Handwriting disabilities criterion.*

| OEE | OEE(t) | HPSQ – C(t) | HDC | No. of children | % |
|---|---|---|---|---|---|
| 0 | 0 | 0 | 0 | | |
| 1 | 0 | 0 | 0 | | |
| 2 | 0 | 0 | 0 | 173 | 49 |
| 2 | 0 | 1 | 0 | | |
| 3 | 1 | 0 | 1 | 108 | 30.59 |
| 3 | 1 | 1 | 2 | 61 | 17.29 |
| 4 | 1 | 0 | 2 | | |
| 4 | 1 | 1 | 3 | 11 | 3.11 |

[a]OEE – overall expert evaluation; OEE(t) – overall expert evaluation (thresholded); HPSQ – C(t) – child's self-evaluation (thresholded); HDC – new handwriting disabilities criterion.

#### 2.4.1. Handwriting/Drawing parameterization

The process of writing/drawing was sampled by the digitizer with sampling frequency $f_s = 133$Hz. The tablet recorded the following time series/signals: x and y position ($x[n]$ and $y[n]$); timestamp ($t[n]$); a binary variable ($b[n]$), being 0 for in-air movement (i. e. movement of pen tip up to 1.5 cm above the tablet's surface (Alonso-Martinez, Faundez-Zanuy, & Mekyska, 2017)) and 1 for on-surface movement (i. e. movement of pen tip on the paper), respectively; pressure exert on the tablet's surface during writing ($p[n]$); pen tilt ($a[n]$); azimuth ($az[n]$). Position is expressed in millimeters and time in seconds.

In the next step, the signals were parameterised employing elementary features that could be split into several groups:

- temporal – duration of writing (DUR), ratio of the on-surface/in-air duration (DURR), duration of strokes (SDUR), and ratio of the on-surface/in-air stroke duration (SDURR).

- kinematic – velocity (VEL), acceleration (ACC), signal-to-noise ratio between the original velocity profile and the one reconstructed by the sigma-lognormal model (SNR) (Duval, Rémi, Plamondon, Vaillant, & O'Reilly, 2015; Ferrer, Diaz, Carmona- Duarte, & Plamondon, 2020), number of lognormal functions required to recon- struct the original velocity profile (nbLog) , and global indicator of the graphomotor performance given as a fraction of SNR and nbLog (SNR/nbLog) (Duval, Rémi, Plamondon, Vaillant, & O'Reilly, 2015; Ferrer, Diaz, Carmona-Duarte, & Plamondon, 2020).

- dynamic – pressure (PRESS), tilt (TILT), and azimuth (AZIM).

- spatial – stroke height (SHEIGHT)

- spiral-specific – degree of spiral drawing severity (DoS) , mean drawing speed of spiral (MDS) , second-order smoothness of spiral (2ndSm) , spiral precision index (SPI) (Cascarano et al., 2019), spiral tightness (TGHTNS) , variability of spiral width (SWVI) , and first-order zero-crossing rate of spiral (1stZC) (San Luciano et al., 2016).

- loops/zig-zag-line/arcade-specific – local minima (LMIN), local maxima (LMAX), distance between neighbour local maxima (DLMAX), velocity at local maxima (VLMAX), width of teeth of the zig-zag task (DFB; on a horizontal line going through 95% of a particular tooth height), normalised width of teeth (NDFB; DFB normalised by a mean distance between local minima), and distance between neighbour bows of the arcade task (DBB; on a horizontal line going through 50% of the first of them))

- other – number of interruptions (pen elevations; NINT), number of pen stops (NPS) (Paz-Villagrán, Danna, & Velay, 2014), tempo (TEMPO; number of strokes normalised by duration), number of on-surface intra-stroke intersections (NIAI), relative number of on-surface intra-stroke intersections (RNIAI), number of on-surface inter- stroke intersections (NIEI), and relative number of on-surface inter-stroke intersections (RNIEI), handwriting density (ADEN) (Asselborn et al., 2018), density on path (PDEN) (Asselborn, Chapatte, & Dillenbourg, 2020), median of power spectral density of speed frequencies (MPSSF) (Asselborn et al., 2018), median of power spectral density of tremor frequencies (MPSTF) (Asselborn et al., 2018), Lempel-Ziv complexity (LZC) (Aboy, Hornero, Abásolo, & Álvarez, 2006; Mekyska et al., 2016), Shannon entropy (SHE) (Mekyska et al., 2016), number of changes in the velocity profile (NCV), relative number of changes in the velocity profile (RNCV), number of changes in the azimuth profile (NCA), number of changes in the pressure profile (NCP), number of changes in the tilt profile (NCT), number of changes in the x profile (NCX), and number of changes in the y profile (NCY).

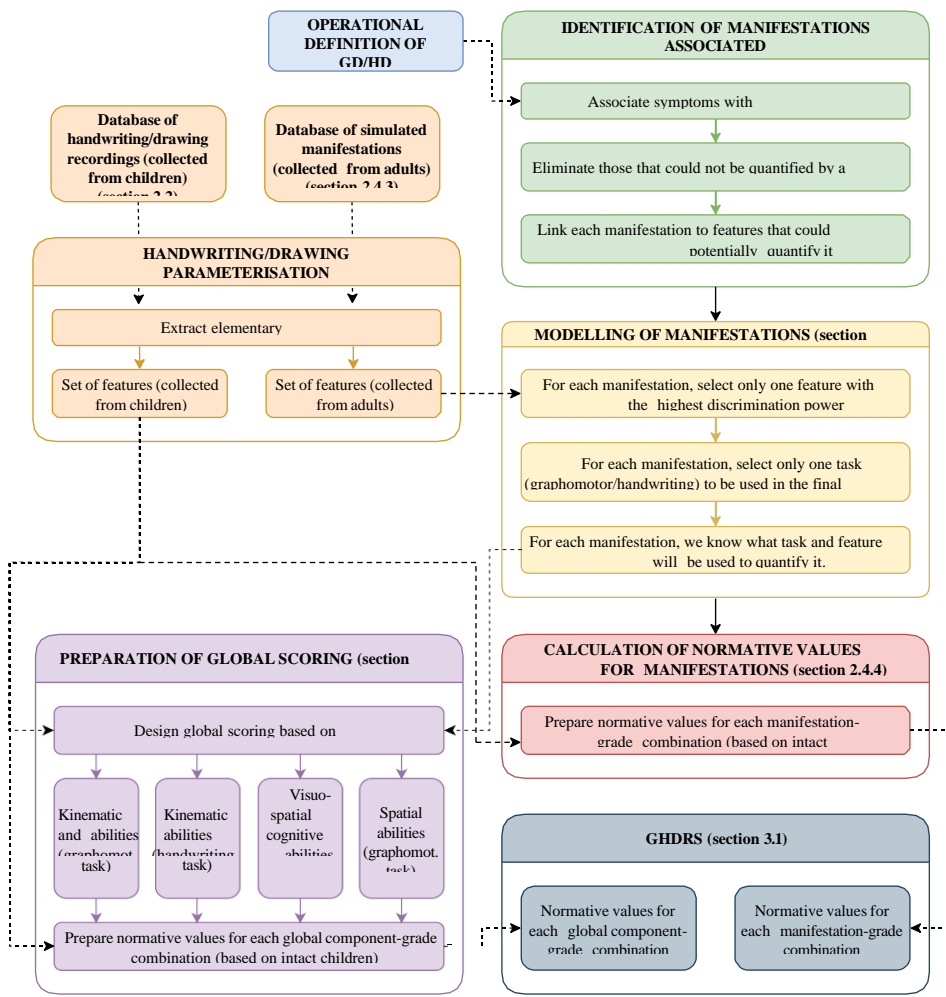

*Figure 3. Design of the GHDRS.*

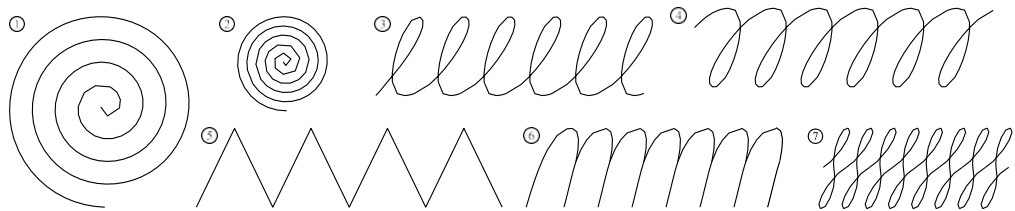

*Figure 4. Graphomotor part of the acquisition protocol.*

The majority of the features were extracted using the freely available Python library handwriting-features (v 1.0.1) (Galaz, Mucha, Zvoncak, & Mekyska, 2022), the rest of them were coded in Matlab (Ferrer, Diaz, Carmona-Duarte, & Plamondon, 2020). Tempo and duration-based features were extracted from both on-surface as well as in-air movements. In addition, some features were analysed in horizontal and vertical projection (see Section 2.4.2). Features represented by a vector (e.g. velocity or local minima) were transformed into a scalar value using median, interquartile range (iqr), non-parametric coefficient of variation (ncv; defined as iqr/median), 95th percentile (95p), and/or slope by applying the Theil – Sen estimator (slope). For each feature, we use the following notation: INF: DIR-FN (HL), where INF stands for processed information (ON for on-surface, AIR for in- air, PRESS for pressure, TILT for tilt, and AZIM for azimuth), DIR denotes direction (H for horizontal and V for vertical), FN contains feature name, and HL a statistic that was used for the transformation. For example, ON: V-VLMAX (median) stands for median of on- surface vertical velocity in local maxima.

2.4.2. Identification of manifestations associated with graphomotor and handwriting disabilities.

Based on the review summarized in the Introduction section we identified a set of symptoms that could be observed in children with GD/HD. As already mentioned, a symptom could be associated with several manifestations. Thus, in the following step, for each symptom, we identified possible manifestations and then eliminated those that could not be assessed, or are difficult to assess by digitizing tablets, e.g. inappropriate shape of letters, a combination of cursive and block letters, etc. Finally, we split the set into subsets associated with the handwriting/drawing process and product, respectively. Each of the remaining manifestations could be quantified utilising a parameterization of signals acquired by the digitizer. Therefore, we heuristically linked the manifestations with features that were good candidates for their quantification (a finer and final selection of features was done during machine-learning-based modelling, see Section 2.4.3). The identified manifestations and their initial pairing with the online handwriting/drawing features could be seen in Tables 5 and 6.

2.4.3. Modelling of manifestations

Since some manifestations were initially linked with several features extracted from several tasks, we had to address two questions:
Q1 What features are the most important ones for quantification of the manifestations?
Q2 Which task mostly accents the manifestation?
To answer Q1 we needed a representative and perfectly labelled dataset of TSK1–10 that contained samples with and without a manifestation. As far as we know, such a dataset does not exist. Moreover, compiling such a dataset would be extremely demanding, because, as already mentioned, two children diagnosed with GD/HD could experience completely different symptoms. Therefore, in order to select an appropriate feature (quantification measure) for each manifestation, we decided to generate a dataset in which the manifestations are simulated under conditions where performance is specifically controlled.

For this purpose, we enrolled 4 proficient handwriters (1 female and 3 males, age 29:3 ±4:0 years) who performed a detailed protocol that is available in the Appendix A6. After that, for each combination of a manifestation/task, we had 20 samples of intact handwriting/drawing and 12 samples corresponding to the specified manifestation.

It was not crucial for the proficient handwriters to perfectly simulate the handwriting/drawing of children with GD/HD. The simulated data were not used to estimate normative values, and they were not used in the subsequent statistical analysis, but only to answer Q1, i.e. to find the most discriminating features. To draw a parallel, let's consider measuring a child's temperature when uncertain whether to use a hand or a thermometer. Using an adult database, we might conclude that the thermometer is more appropriate. However, the final measurement using the thermometer is performed on the child.

To select the most appropriate features, we employed simple logistic regression with $\ell 1$ regularization (LASSO) with the following parameters (parameters that are not mentioned in the list were fine-tuned or kept using the default settings of the LogisticRegression class that is part of the scikit-learn 0.24.2 library): penalty = $\ell 1$; class weight = balanced; solver = liblinear; random seed = 42.For each manifestation/task combination, we:

1)Standardized the features on a per-feature basis to have the mean of 0 and the standard deviation of 1.

2)Optimised the hyper-parameters utilising the grid-search strategy. We fine-tuned the regularization parameter (C) of the LASSO regression using the grid of 250 logarithmically distributed values from log(-2) to log(0:25). For this purpose, we employed the leave-one-group-out cross-validation, i. e. in each cross-validation fold, we trained the model with all but one group (proficient handwriters) and tested it by the remaining one (the remaining proficient handwriter). The performance of the trained models was evaluated by balanced accuracy (BACC) score that represents the arithmetic mean between the model's sensitivity and specificity.

3)Trained the final model with the fine-tuned set of hyper-parameters.

Even though a manifestation in one task could be theoretically modelled by a linear combination of several features (depending on the ,1 regularization), to make the 16J. MEKYSKA ET AL.

15*Table 5.* Manifestations associated with the process of handwriting/drawing and proposed sets of features quantifying them. Each manifestation is paired with a symptom (see table 1).

| Manifestation [Symptom] | Task | Feature | Feature definition |
|---|---|---|---|
| Higher duration of writing [Variability in speed] | TSK8–10 | DUR | overall duration |
| | | ON: DUR | duration of on-surface movement |
| | | ON: SDUR (median) | median duration of on-surface strokes |
| Low velocity [Variability in speed] | TSK1–10 | ON: G,H,V-VEL (median) | median of on-surface global/horizontal/vertical velocity |
| | | ON: G,H,V-VEL (95p) | 95th percentile of on-surface global/horizontal/vertical velocity |
| | TSK1–2 | MDS | mean drawing speed of spiral |
| Low acceleration [Variability in speed] | TSK3–10 | ON: G,H,V-ACC (median) | median of on-surface global/horizontal/vertical acceleration |
| | | ON: G,H,V-ACC (95p) | 95th percentile of on-surface global/horizontal/vertical acceleration |
| Lower variability of velocity [Variability in speed] | TSK8–10 | ON: G,H,V-VEL (iqr) | interquartile range of on-surface global/horizontal/vertical velocity |
| Lower variability of acceleration [Variability in speed] | TSK8–10 | ON: G,H,V-ACC (iqr) | interquartile range of on-surface global/horizontal/vertical acceleration |
| Dysfluency in velocity [Variability in speed] | TSK1–7 | ON: NCV | number of changes in on-surface velocity profile |
| | | ON: RNCV | relative number of changes in on-surface velocity profile |
| | | ON: MPSSF | median of power spectral density of speed frequencies |
| | | ON: SNR | signal-to-noise ratio between original velocity profile and the one reconstructed by sigma-lognormal model |
| | | ON: nbLog | number of lognormal functions required to reconstruct original velocity profile |
| | | ON: SNR/nbLog | global indicator of graphomotor performance given as a fraction of SNR and nbLog |
| | | ON: NPS | number of pen stops |
| Gradually decreasing velocity [Unusual wrist and body position] | TSK8–10 | ON: G,H,V-VEL (slope) | slope of on-surface global/horizontal/vertical velocity |
| Gradually decreasing acceleration [Unusual wrist and body position] | TSK8–10 | ON: G,H,V-ACC (slope) | slope of on-surface global/horizontal/vertical acceleration |
| Unstable pressure on pen tip [Unstable pressure] | TSK1–10 | PRESS (ncv) | non-parametric coefficient of variation of pressure |
| | | PRESS (slope) | slope of pressure profile |
| | | NCP | number of changes in pressure profile |
| Unstable tilt of pen [Unstable tilt] | TSK1–10 | TILT (ncv) | non-parametric coefficient of variation of tilt |
| | | NCT | number of changes in tilt profile |
| Visuospatial deficits [Longer in-air time period] | TSK8–10 | AIR: DUR | duration of in-air movement |
| | | AIR: SDUR (median) | median duration of in-air strokes |
| | | DURR | ratio of on-surface/in-air duration |
| Disability to perform longer strokes (frequent interruptions and pen elevations) [Dysfluency of handwriting] | TSK8–10 | NINT | number of interruptions |

*Table 6. Manifestations associated with handwriting/drawing products and proposed sets of features quantifying them. Each manifestation is paired with a symptom (see table 1).*

| Manifestation [Symptom] | Task | Feature | Feature definition |
|---|---|---|---|
| Instability in amplitude of letters [Problems with size control] | TSK3–5 | ON: V-LMAX (ncv) | non-parametric coefficient of variation of local maxima in vertical projection |
| | TSK8–10 | ON: SHEIGHT (ncv) | non-parametric coefficient of variation of stroke height |
| All letters have same amplitude [Problems with size control] | TSK8–10 | ON: SHEIGHT (ncv) | non-parametric coefficient of variation of stroke height |
| Inability to maintain handwriting on a line [Messy organization] | TSK3, 5, 6 | ON: V – LMIN (ncv) | non-parametric coefficient of variation of local minima in vertical projection |
| Frequent overwriting [Grammar mistakes] | TSK8–10 | ON: NIEI | number of on-surface inter-stroke intersections |
| | | ON: RNIEI | relative number of on-surface inter-stroke intersections |
| Unstable density [Poor spacing] | TSK1–10 | ON: PDEN | density of path |
| | TSK1–10 | ON: ADEN | handwriting density |
| | TSK3–4 | ON: V – DLMAX (ncv) | non-parametric coefficient of variation of distance between neighbour local maxima in vertical projection |
| | TSK1–2 | SPI | spiral precision index |
| | TSK1–2 | TGHTNS | spiral tightness |
| | TSK1–2 | SWVI | variability of spiral width |
| | TSK1–4, 7–10 | ON: NIAI (median) | median of number of on-surface intra-stroke intersections |
| | | ON: RNIAI (median) | median relative number of on-surface intra-stroke intersections |

whole process simple, we decided to always use one feature for each manifestation. To do so, we took all combinations modelling a manifestation (e.g. TSK1–7 for dysfluency in velocity), sorted features in each LASSO model (7 models for TSK1–7) by weights and selected the feature that appeared to be the most important across all the models (e.g. the number of lognormal functions – nbLog). In addition, using the database of children, we ensured that the higher values for a feature meant worse performance. Otherwise, e.g. in the median of velocity representing the manifestation of low velocity, we multiplied the value by 1. At the end of this step, we answered question Q1, i.e. for each manifestation we identified the most discriminative feature.

To answer the Q2 and pair a manifestation with the most appropriate task, we assumed that in the intact cohort, the manifestation should be "softened" with increasing grade. e.g. the dysfluency in velocity should decrease going from the pre-schoolers until the 4th-grade children. Therefore, we plotted a scatter diagram and chose the task with decreasing monotonic trend and with a high Spearman's rank correlation coefficient. At the end of this step, we paired one manifestation with one feature extracted from one task

2.4.4. Calculation of normative values for manifestations

For each grade, we prepared normative values for the manifestations. However, some manifestations could not be assessed in early grades, e.g. in children who are not able to write. To prepare the normative values, we followed these steps:

- For each manifestation m and each child i, we calculated a specific feature $f_m^i$ from a specific task (see Section 2.4.3). Variable i=1: $I$, where $I$ is the number of children in the specific grade.

- We performed min-max scaling so that feature values (f i for all i) were in the range from 0 to 1.

- Using just the intact cohort, we calculated the median ($f_m$) and a threshold, which was initially set to be the 95th percentile of $f_m$.

- We plotted an estimation of probability density function of $f_m^i$ (for all $i$) using the Gaussian kernel (see function ksdensity of Matlab R2021a). In the same graph, we marked intact children, as well as the ones diagnosed with DD. If necessary, we empirically adjusted the threshold so that it was able to distinguish children with and without GD/HD.

Such an approach is capable of both:

(1) identification of a manifestation for each child – we consider that a manifestation is identified if the associated feature is greater than the threshold,
(2) rating its severity – severity is rated by a score calculated using the formula

$$s_m^i = \frac{f_m^i - \text{median}}{\text{threshold} - \text{median}}, \quad (1)$$

i.e. if $f_m^i$ = threshold then $s_m^i$=1; a manifestation/symptom is more severe with increasing $s_m^i$.

During the design of the GHDRS, we paid much attention to its easy integration. Therefore, for each grade and manifestation separately we provide the community with all information necessary for diagnosis/rating, i.e. with the name of the feature, corresponding weight (1 or −1), minimum, maximum (necessary for the min-max scaling); and normative values – median and threshold.

In order to explore whether we can group some features $f_m$ and get a more global overview of the child's performance, we employed principal component analysis (PCA) with the promax rotation method, and parallel analysis to determine the optimal number of components (the analysis was done in JASP, version 0.16.2 (JASP Team, 2022)). The PCA was fed by the raw values of the features ($f_m$, see section 2.4.4) and noy by the score $s_m^i$.. We introduced global features based on these steps:

1) Number of global features $o_j$ (j=1:J) is equal to the number of component, i.e.
2) A global feature is given as a linear combination $o_j = \sum_m W_m \cdot f_m$, where $f_m$, are features whose loadings in a particular component were $\geq 0.4$, $w$ are weights derived from loadings, but normalized so that $\sum_m W_m = 1$. As with a particular score, a global score could be calculated as:

$$g_j^i = \frac{o_j^i - \text{median}}{\text{threshold} - \text{median}}, \qquad (2)$$

where the median and threshold are again calculated from the data of the intact children. All information necessary for the diagnosis/rating is provided.

## 3. Results

### 3.1. The GHDRS

Following the process described in Section 2.4.3 and 2.4.4, for each manifestation we identified an optimal combination of a task and feature that enables computerized assessment. These combinations are reported in Appendix A7 – Table 1 (grades 0–1), Appendix A7 – Table 2 (grades 2–3) and Appendix A7 – Table 3 (grade 4). In addition, for each grade and manifestation separately, we provide feature weight; minimum and maximum (necessary for the min-max scaling); and normative values, i.e. the median and threshold. In other words, the tables enable the deployment of the GHDRS.

Results of the PCA (see Section 2.4.5) are reported in Appendix A7 – Table 4. After the analysis of the results, we decided to group the manifestations (more specifically the associated features) into four global components (corresponding weights and normative values could be found in Appendix A7 – Table 5):

C1 → global component G1 – Kinematic abilities (graphomotor task)
C2 → global component G2 – Kinematic abilities (handwriting task)
C3 → global component G3 – Visuo-spatial and cognitive abilities (handwriting task)
C4 → global component G4 – Spatial abilities (graphomotor task)

The first global component assesses Kinematic abilities (based on velocity and acceleration) when performing the graphomotor task (i.e. it quantifies the process). The fourth component focuses on the product of the task (e.g. whether a child keeps consistent spacing between neighbour loops or whether s/he maintains the same amplitude of the loops). The second and third components quantify handwriting abilities. More specifically, the second one quantifies Kinematic abilities (process), e.g. whether the child maintains the same velocity, and the third is a combination of Visuo-spatial and cognitive abilities (mainly linked with the product), e.g. whether s/he avoids frequent overwriting. For simplified referencing, in the following text, we will call G1 and G2 as kinematic component, and G3 and G4 as product ones.

*3.1. GHDRS psychometric properties*

Shapiro–Wilk tests showed that scores on Kinematic abilities (graphomotor task) ($W_{334} = 0.96$, p < .001), Spatial abilities (graphomotor task) ($W_{334} = 0.82$, p < .001), Kinematic abilities (handwriting task) ($W_{185} = 0.95$, p < .001) and Visuo-spatial and cognitive abilities (handwriting task) ($W_{185} = 0.97$, p < .001) had distributions that departed significantly from normality. Values of skewness and kurtosis were within acceptable limits of 2 (Field, 2013; Gravetter, Wallnau, Forzano, & Witnauer, 2020; Trochim & Donnelly, 2001) except for Spatial abilities (graphomotor task), where skewness reached a value of 2.13 and kurtosis of 6.94, and Visuo-spatial and cognitive abilities (handwriting task), where the value of kurtosis was 2.68. Both kinematic scores are left-skewed, and both product scores are right-skewed. Based on this outcome, a non-parametric test was used for analysis. To analyze the data, we used JASP Team (JASP Team, 2022).

Differences in sex were tested with an alternative hypothesis specification that the scores of boys were not equal to that of girls. Levene's homogeneity tests were non-significant, except for Spatial abilities (graphomotor task) ($p = .03$). Kinematic abilities score (graphomotor task) did not differentiate between sexes ($U = 13646.5$; $p = .89$; $r_B = -0.01$). A Welch test found a significant difference for Spatial abilities with a large effect size ($W(331.97) = 4.26$; p < .001; g = 0.46), where boys (N = 186; M = 0.6; SD = 1.4) performed worse than girls (N = 148; M = −0.01; SD = 1.1). Similarly, there was a significant sex difference from a Mann–Whitney test of the Kinematic abilities (handwriting task a with small effect size (U = 3341; p = .02; rB = −0.20) with higher scores for girls (N = 79; M= 0.1; SD = 0.7) than boys (N = 106; M = −0.2; SD = 0.9). Lastly, Visuo-spatial and cognitive abilities differentiated between the two groups with medium effect ($U = 5113$; $p < .001$; $r_B = 0.22$), with boys ($N = 106$; $M = 0.2$; $SD = 0.7$) performing worse than girls ($N = 79$; $M = -0.1$; $SD = 0.7$).

Pairwise correlations were performed to explore relationships. A positive correlation was found between Kinematic abilities (graphomotor and handwriting tasks: $\rho = .45$, p< .001) and Spatial abilities (graphomotor and handwriting tasks: ρ = .27, p < .001). A weak negative relationship were observed between the Kinematic abilities (handwriting task) and the Spatial abilities (graphomotor task) ($\rho = -.17$, $p = .02$) and the Visuo-spatial and cognitive abilities (handwriting task) ($\rho = -.16, p = .03$). No significant correlations were found between the Kinematic abilities (graphomotor task) and the Spatial abilities (graphomotor task) ($\rho = .04, p = .52$), and Visuo-spatial and cognitive abilities ($\rho = -.04, p = .62$).

A separate expert evaluation of the upper loops (TSK3) was used to explore the validity of both graphomotor ability scores. Analogously, a separate expert assessment of the transcription task (TSK8–10) was used to validate both handwriting ability scores. The alternative hypothesis was set for positive correlation. Spearman correlation found a significant relationship between expert evaluation and both graphomotor scores, for Kinematic abilities ($\rho = .41$; $p < .001$) and Spatial abilities ($\rho = .22$; $p < .001$). Similar outcomes were found for the Visuo-spatial and cognitive abilities ($\rho = .23$; $p < .001$). On the other hand, Kinematic abilities (handwriting task) did not significantly correlate with expert assessment ($\rho = -.15; p = .98$).

In the end, the Handwriting legibility scale (Barnett, Prunty, & Rosenblum, 2018) and a shortened version of the Concise Assessment Methods of Children's Handwriting scale (SOS2-EN) (Van Waelvelde, Hellinckx, Peersman, & Smits-Engelsman, 2012) were employed to test the construct validity of both handwriting scores. Using the Spearman correlation coefficient, the alternative hypothesis was set as a positive correlation, where higher scores mean worse performance. The Kinematic abilities score did not significantly correlate with HLS ($\rho = -.24$; $p = .93$) nor with SOS: BHK ($\rho = -.16$; $p = .95$). On the other hand, the Visuo-spatial and cognitive score showed moderate to strong significant correlations with HLS ($\rho = .48$; $p < .001$) and BHK ($\rho = .58$; $p < .001$).

## Discussion

GHDRS is the first scale enabling objective and interpretable assessment of manifestations associated with GD/HD. The scale could be used to quantify these manifestations in preschool children, who are still learning how to write, as well as in children who should have already mastered handwriting, while its global assessment give an overall profile of the product/process of drawing/handwriting. The graphomotor part of the acquisition protocol consisted of 7 tasks (see Figure 4), and we observed that in most cases the model correctly quantified manifestations from the TSK3, i.e. upper loops (sometimes called a spring), which is commonly used for screening GD (Galaz et al., 2020; Galli et al., 2011; Meulenbroek & Van Galen, 1986; Vaivre-Douret, Lopez, Dutruel, & Vaivre, 2021). Therefore, normative data about the task can be derived using this procedure which also significantly simplifies the whole assessment process, since only the graphomotor element and, in children who know how to write, the transcription, are required for the final protocol. Prior to this, probably the only (pioneering) work dealing with a complex data-driven based assessment of HD was published by the team of Asselborn et al. (Asselborn, Chapatte, & Dillenbourg, 2020). In contrast, other attempts have focused on modeling/ simulating the ratings performed by a human (often using a scale) (Asselborn et al., 2018; Devillaine et al., 2021; Galaz et al., 2020; Mekyska et al., 2016, 2019; Rosenblum & Dror, 2016), i.e. the effect of subjectivity could have played some role. But Asselborn et al. employed a data-driven approach and introduced a new methodology to HD assessment based on one global score and four sub-scores, namely: 1) kinematic, 2) pressure, 3) tilt, and 4) static. If we compare the outcomes of their study to the results we report here, then their sub-scores are on the same level as our global scores. More specifically, we use two global scores to analyse handwriting, G2 – kinematic abilities (which is equivalent to the kinematic and tilt sub-scores of Asselborn et al.) and G3 – visuo-spatial and cognitive abilities (which is equal to the pressure and static sub-scores). We went further and also linked the input features with specific manifestations, i.e. GHDRS facilitates interpretation and understanding of specific HD. Moreover, we extended the assessment to GD as well, thus enabling the examination of pre-school children.

Previous studies have observed that boys have worse handwriting quality (Hawke, Olson, Willcut, Wadsworth, & DeFries, 2009; Šafárová et al., 2020). Results from GHDRS imply the same pattern since both product scores distinguish between boys and girls, with boys having a worse performance. Research outcomes for handwriting speed are ambiguous.

The Kinematic abilities score (graphomotor task) did not show any differences between sexes, which corroborates other recent findings (Wicki, Lichtsteiner, Geiger, & Müller, 2014). On the other hand, the surprising result brings the Kinematic abilities score (handwriting task) with slower handwriting in girls, corresponding to Van Galen's study (van Galen, 1991). Although the Kinematic abilities score (graphomotor task) did not yield a significant difference, girls were slower than boys, corresponding with the results of the other kinematic score. We suspect that these results could be explained by girls possibly being more meticulous in their handwriting.

Thus, a hypothesis about individual differences in performing the task should be considered in future research.

We obtained evidence of a positive relationship between both kinematic scores, which means that children with faster performance in the graphomotor task have faster performance in the handwriting task. Both product scores work analogically. Children with worse graphomotor drawing have poorer handwritten outcome. The negative relationship of the Kinematic abilities score (handwriting task) to the product score is interesting. It indicates that a lower handwriting speed is related to a neater product. These results contribute to the discussion about the speed-legibility trade-off (Karlsdottir & Stefansson, 2002; Weintraub & Graham, 1998) and could be supported by previous studies. Blote et al. (Blöte & Hamstra-Bletz, 1991) described the non-linear relation between legibility and speed of handwriting. They found that children in higher grades with poor handwriting write either slowly or fast. Other studies found that speed does not degrade legibility (Graham, Weintraub, & Berninger, 1998; Wicki, Lichtsteiner, Geiger, & Müller, 2014) or that there are no significant relationships (Rubin & Henderson, 1982). So far, we might reasonably say that children with slower handwriting do not necessarily have poor handwriting (Kushki, Schwellnus, Ilyas, & Chau, 2011).

Lastly, our findings indicate that being evaluated more negatively by an expert coincided with worse performance on all scales except the Kinematic score (handwriting task). Correspondingly, the poorer performance on the HLS (Barnett, Prunty, & Rosenblum, 2018) and SOS: BHK (Van Waelvelde, Hellinckx, Peersman, & Smits-Engelsman, 2012) was related to worse performance on the Visuo-spatial and cognitive score, but not the Kinematic score. Moreover, despite the absence of a significant relationship, slower handwriting had inverse correlations with product evaluation of expert, SOS: BHK and HLS. Several thing might account for this. First, the expert, HLS, and SOS: BHK scales assess only the static product (loops and transcription) and not the handwriting speed or the process of handwriting. A related possibility is that experts, when making their assessment decisions, attach very different weights to the observed features compared to the data-driven approach presented here. For example, evi- dence of overwriting could impact the overall expert's evaluation, whereas the data-driven approach did not ascribe as much importance to this manifestation. Therefore, both sources of information should be used in the diagnostic decision-making process.

Additionally, the nature of the task should be considered. For example, Parush et al. (Parush, Lifshitz, Yochman, & Weintraub, 2010) stated that spatial organization is the best predictor of legibility in copying tasks, whereas speed was significant for dictation tasks. In addition, there are speed/velocity differences in the literature (see variability in velocity in the Introduction section). According to Feder and Majnemer (Feder & Majnemer, 2007), handwriting speed varies depending on the context, instruction and the nature of the task (transcription, dictation, free writing). Therefore, the graphomotor task, which is easier to draw, provides more relevant diagnostic results than the complex transcription task. The distribution of all scores is related to this problem. Both Kinematic scores are left-skewed, which means that the children were performing poorly overall, which could be a reason for the decreased ability to distinguish between children with typical hand-writing development and GD/HD. In contrast, both product scores are right-skewed, which better detects poor drawing or handwriting performance.

These outcomes raise new questions about whether kinematic performance could be used as an independent indicator of GD/HD. Perhaps there exists the possibility of new and more complex definitions for text features describing the manifestation of HD.

Asselborn et al. proved that four children, all diagnosed with DD, could actually have different difficulties (the Asselborn's team call it handwriting profiles) (Asselborn, Chapatte, & Dillenbourg, 2020). One child could manifest impaired kinematic abilities (the process of handwriting) while yields an impaired product. We have observed similar cases in our data. Figure 5 shows the GHDRS of three children attending 3rd grade. As can be seen, the first girl has no GD/HD (intact). The second girl has GD/HD, more specifically, she has the impaired process of drawing and handwriting. Conversely, the boy has impaired product of drawing and handwriting, which could be also seen in Figure 6 (he was not able to maintain the loops in a line and was not able to keep a stable tilt – this probably explains the different orientation of loops) and in Figure 7 (frequent overwriting, disability to perform longer strokes, all letters tended to have the same amplitude).

The present research contributes to a new data-driven approach to diagnosing GD/HD. We want to present a new point of view in this area. Based on our data, we can detect nuances in GD/HD, which could lead to the identification of unique sets of manifestations for each child. GD/HD can not be diagnosed as a sum of symptoms but as a combination of different manifestations with different severity. This view more precisely reflects the incongruities in research studies (e.g. whether children with GH/HD are faster, slower, or show no differences), and the wide range of prevalence (from 7–34%, depending on the assessment criterion). Evidently, children with GD/HD can use different compensation mechanisms and successfully cover up their issues. On the other hand children with typical development could manifest some handwriting issues, which can subside with time. Therefore, a complex description of symptoms with the corresponding manifestations allows experts in the field to better target remediation. In our opinion, this approach better reflects the situation in the field of specific learning disorders.

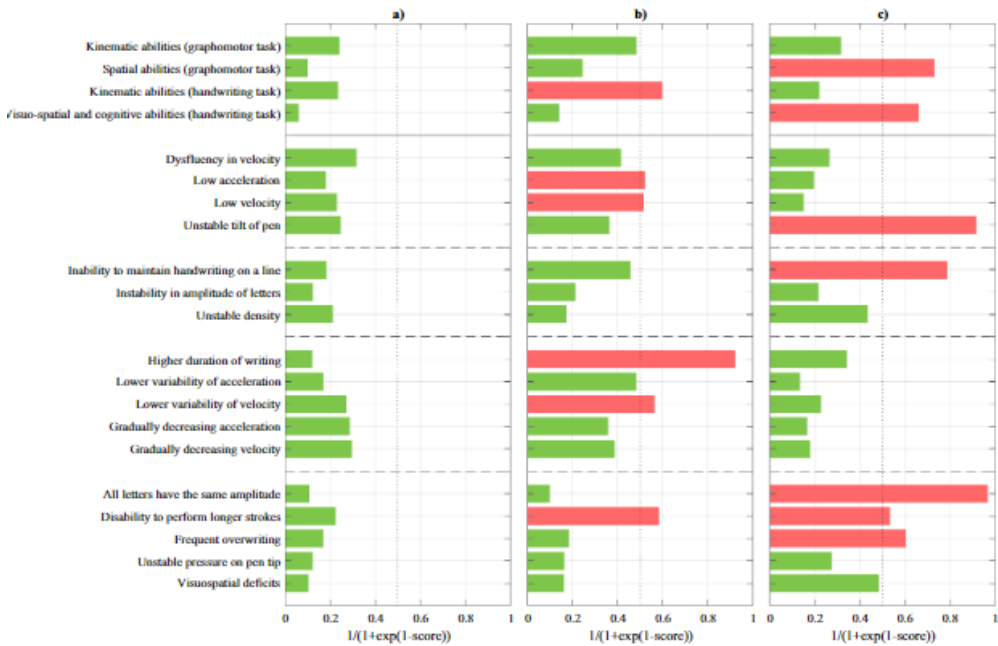

*Figure 5. Three children attending 3rd grade of a primary school assessed based on the GHDRS (the first top block contains the global scores; the next four blocks contain specific manifestations, i.e. they represent a detailed profile associated with the global scores; all scores are transformed by a sigmoid function so that the minimum is 0, maximum 1 and the threshold determining disability has value 0.5): a) an intact girl without any GD/HD; b) a girl with the impairment in the process of handwriting (too high duration of writing, lower variability of velocity) and in the process of drawing loops (low velocity, low acceleration); she is also not able to perform longer strokes during writing; c) a boy whose handwriting is characterized by frequent overwriting (see figure 7), disability to perform longer strokes, moreover, all letters tended to have the same amplitude; in addition, he was unable to maintain loops in a line (see figure 6) and was not able to keep a stable tilt.24J. MEKYSKA ET AL.*

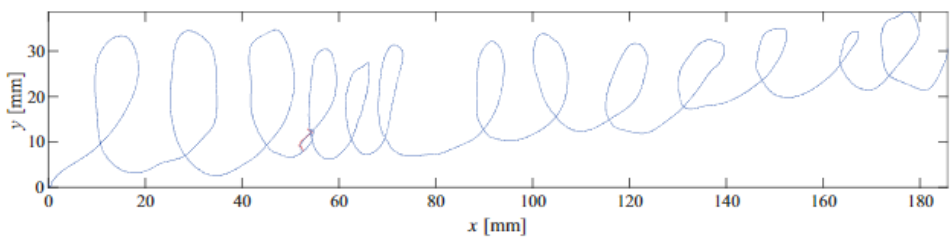

*Figure 6. TSK3 (upper loops) performed by a boy, whose GHDRS is depicted in figure 5c (the blue line represents on-surface movement, the red line represents in-air movement).*

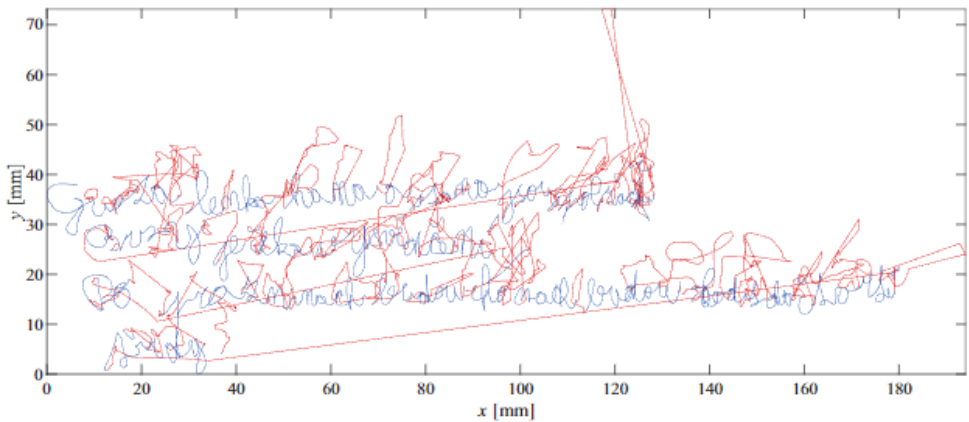

*Figure 7. TSK10 (a paragraph) written by a boy, whose GHDRS is depicted in figure 5c (the blue line represents the on-surface movement, the red line represents the in-air movement).*

### 4.1. Limitations

As reported in the introduction, although there have been several attempts to introduce a tool that could be used to objectively assess graphomotor and handwriting disabilities, this field still contains many knowledge gaps that we tried to bridge with a unique and original approach. On 020406080100120140160180x[mm]0102030y[mm]Figure 6. TSK3 (upper loops) performed by a boy, whose GHDRS is depicted in figure 5c (the blue line represents on-surface movement, the red line represents in-air movement).020406080100120140160180x [mm]010203040506070y[mm]Figure 7. TSK10 (a paragraph) written by a boy, whose GHDRS is depicted in figure 5c (the blue line represents the on-surface movement, the red line represents the in-air movement).

The other hand, since it is unique, we admit that our design has several limitations that could be further addressed:

● small-sample and imbalanced database – As reported in Table 2, our database is relatively small and imbalanced with respect to grade levels. This could negatively affect estimation of normative data. Data collection in pre-school children and those attending up to the fourth grade can be very challenging. We analyzed 353 children, but in fact we enrolled more than 450. Due to several issues (e.g. inability to complete some tasks, missing demographic data, etc.) the sample size was reduced. Nevertheless, we have started to collect new data, and in the next couple of years we plan to update the normative values.

● limited set of manifestations – We modelled 17 manifestations of GD/HD, but we believe it is possible identify more and make the assessment even more complex. We identified these manifestations based on the literature review and on discussions with psychologists and remedial teachers. We tried to quantify the most frequent ones and also the ones that could be reasonably processed by a computer. We are aware that we have left out some manifestations associated with the product of drawing/handwriting (e.g. improper shape of letters), but this requires further research and the introduction of a robust technical solution (e.g. based on convolutional neural networks).

● simple simulation of manifestations – As mentioned in Section 2.4.3, in order to identify an optimal combination of manifestation/task/feature, we decided to simulate the manifestations by four proficient handwriters who always produced 20 samples of intact handwriting/drawing and 12 samples corresponding to the specific manifestation. Although the number of proficient handwriters could have been higher, we used this database to get some initial intuitions about proper combinations, and it was not used to derive the normative data (for this purpose, we used the database of children).

● handwriting disabilities criterion – In order to calculate the normative data just from intact children, we needed to stratify the database using a criterion explained in Section 2.3. This could hypothetically introduce some effect of subjectivity. On the other hand, the criterion is calculated from several sources, and we tried to make it as reliable as possible. Besides, as explained in Section 2.4.4, we did not fully rely on this criterion and, if necessary (e.g. in the case of outliers), adjusted the threshold based on the visual inspection of the probability density function calculated from manifestation scores.

To sum up, even though our study and the scale have several limitations, the study represents an important contribution to the objective and detailed assessment of GD/HD, which could have a positive impact on the research community, but most importantly on children and their quality of life. Because it is a novel instrument, we will further optimize it, adapt it to other languages, and release updated versions. In this regard, we would welcome opinions from other researchers, proposals for other manifestations, and cooperation during adaptation to other languages.

## 5. Conclusion

his work introduces a new scale facilitating objective, complex, and automatic assessment of GD/HD. It provides detailed quantification of 17 manifestations associated with the process/product of drawing/handwriting and enables better-targeted therapy and remediation. The current normative data could be used in a cohort of Czech children (writing with cursive letters) attending up to the fourth grade.

The scale is a product of a multidisciplinary team consisting of psychologists, psychometricians, remedial teachers, signal processing engineers and data scientists. The whole team paid special attention to its easy integration into practice; therefore, the methodology is described in detail so that the framework and the concept of the scale could be adapted into other languages.

26## References

- ReferencesAboy, M., Hornero, R., Abásolo, D., & Álvarez, D. (2006). Interpretation of the Lempel-Ziv complexity measure in the context of biomedical signal analysis. IEEE Transactions on Biomedical Engineering, 53(11), 2282–2288. doi:10.1109/TBME.2006.883696
- Alonso-Martinez, C., Faundez-Zanuy, M., & Mekyska, J. (2017). A comparative study of in-air trajec-tories at short and long distances in online handwriting. Cognitive Computation, 9(5), 712–720. doi:10.1007/s12559-017-9501-5
- American Psychiatric Association. (2013). Diagnostic and statistical manual of mental disorders. Washington, DC.Asselborn, T., Chapatte, M., & Dillenbourg, P. (2020). Extending the spectrum of dysgraphia: A data driven strategy to estimate handwriting quality. Scientific Reports, 10(1), 1–11. doi:10.1038/ s41598-020-60011-8
- Asselborn, T., Gargot, T., Kidziński, Ł., Johal, W., Cohen, D., Jolly, C., & Dillenbourg, P. (2018). Automated human-level diagnosis of dysgraphia using a consumer tablet. NPJ Digital Medicine, 1(1), 1–9. doi:10.1038/s41746-018-0049-x
- Barnett, A. L., Prunty, M., & Rosenblum, S. (2018). Development of the handwriting legibility scale (hls): A preliminary examination of reliability and validity. Research in Developmental Disabilities, 72, 240–247. doi:10.1016/j.ridd.2017.11.013   Berninger, V. W., & Amtmann, D. (2003). Preventing written expression disabilities through early and continuing assessment and intervention for handwriting and/or spelling problems: Research into practice. In H. L. Swanson, K. R. Harris, & S. Graham (Eds.), Handbook of learning disabilities (pp. 345–363). The Guilford Press.Biotteau, M., Danna, J.,
- Baudou, É., Puyjarinet, F., Velay, J.-L., Albaret, J.-M., & Chaix, Y. (2019). Developmental coordination disorder and dysgraphia: Signs and symptoms, diagnosis, and rehabilitation. Neuropsychiatric Disease and Treatment, 15, 1873. doi:10.2147/NDT.S120514
-Blöte, A. W., & Hamstra-Bletz, L. (1991). A longitudinal study on the structure of handwriting. Perceptual and Motor Skills, 72(3), 983–994. doi:10.2466/pms.1991.72.3.983
- Bosga-Stork, I. M., Bosga, J., & Meulenbroek, R. G. (2017, apr). Emerging behavioral flexibility in loop writing: A longitudinal study in 7- to 9-year-old primary school children. Motor Control, 21(2), 195–210. doi:10.1123/mc.2015-0077
- Burton, A. W., & Dancisak, M. J. (2000, jan). Grip form and graphomotor control in preschool children. The American Journal of Occupational Therapy, 54(1), 9–17. doi:10.5014/ajot.54.1.9
- Cascarano, G. D., Loconsole, C., Brunetti, A., Lattarulo, A., Buongiorno, D., Losavio, G. & Bevilacqua, V. (2019). Biometric handwriting analysis to support Parkinson's disease assessment and grading. BMC Medical Informatics and Decision Making, 19(S9), 1–11. doi:10.1186/s12911-019-0989-3
- Cermak, S. A., & Bissell, J. (2014, may). Content and construct validity of here's how i write (HHIW): A child's self-assessment and goal setting tool. The American Journal of Occupational Therapy, 68 (3), 296–306. doi:10.5014/ajot.2014.010637

- Chang, S.-H., & Yu, N.-Y. (2013). Handwriting movement analyses comparing first and second graders with normal or dysgraphic characteristics. Research in Developmental Disabilities, 34(9), 2433–2441. doi:10.1016/j.ridd.2013.02.028
- Chung, P. J., Patel, D. R., & Nizami, I. (2020, feb). Disorder of written expression and dysgraphia: Definition, diagnosis, and management. Translational Pediatrics, 9(S1), S46–S54. doi:10.21037/tp. 2019.11.01
- Cornhill, H., & Case-Smith, J. (1996, oct). Factors that relate to good and poor handwriting. The American Journal of Occupational Therapy, 50(9), 732–739. doi:10.5014/ajot.50.9.732
- Devillaine, L., Lambert, R., Boutet, J., Aloui, S., Brault, V., Jolly, C., & Labyt, E. (2021). Analysis of graphomotor tests with machine learning algorithms for an early and universal prediagnosis of dysgraphia. Sensors, 21(21), 7026. doi:10.3390/s21217026
- Döhla, D., & Heim, S. (2016, jan). Developmental dyslexia and dysgraphia: What can we learn from the one about the other? Frontiers in Psychology 6. 10.3389/fpsyg.2015.02045 .Drotár, P., & Dobeš, M. (2020). -Dysgraphia detection through machine learning. Scientific Reports, 10 (1), 1–11. doi:10.1038/s41598-020-78611-9 28J.
- Duval, T., Rémi, C., Plamondon, R., Vaillant, J., & O'Reilly, C. (2015). Combining sigma- lognormal modeling and classical features for analyzing graphomotor performances in kindergarten children. Human Movement Science, 43, 183–200. doi:10.1016/j.humov.2015.04.005
- Engel-Yeger, B., Nagauker-Yanuv, L., & Rosenblum, S. (2009, mar). Handwriting performance, self-reports, and perceived self-efficacy among children with dysgraphia. The American Journal of Occupational Therapy, 63(2), 182–192. doi:10.5014/ajot.63.2.182
- Falk, T. H., Tam, C., Schellnus, H., & Chau, T. (2011). On the development of a computer-based handwriting assessment tool to objectively quantify handwriting proficiency in children. Computer Methods and Programs in Biomedicine, 104(3), e102–e111. doi:10.1016/j.cmpb.2010. 12.010
- Faundez-Zanuy, M., Mekyska, J., & Impedovo, D. (2021). Online handwriting, signature and touch dynamics: Tasks and potential applications in the field of security and health. Cognitive Computation, 13(5), 1406–1421. doi:10.1007/s12559-021-09938-2 Feder, K. P., & Majnemer, A. (2003). Children's handwriting evaluation tools and their psychometric properties. Physical and Occupational Therapy in Pediatrics, 23(3), 65–84. doi:10.1080/ J006v23n03_05
- Feder, K. P., & Majnemer, A. (2007, apr). Handwriting development, competency, and intervention. Developmental Medicine & Child Neurology, 49(4), 312–317. doi:10.1111/j.1469-8749.2007.00312.x
- Ferrer, M. A., Diaz, M., Carmona-Duarte, C., & Plamondon, R. (2020). iDelog: Iterative dual spatial and kinematic extraction of sigma-lognormal parameters. IEEE Transactions on Pattern Analysis and Machine Intelligence, 42(1), 114–125. doi:10.1109/TPAMI.2018.2879312
- Field, A. (2013). Discovering statistics using ibm spss statistics. London, UK: sage.

- Flower, L., & Hayes, J. R. (1981, dec). A cognitive process theory of writing. College Composition and Communication, 32(4), 365. doi:10.2307/356600
- Galaz, Z., Mucha, J., Zvoncak, V., & Mekyska, J. (2022). Handwriting Features. https://github.com/ BDALab/handwriting-features. GitHub.
- Galaz, Z., Mucha, J., Zvoncak, V., Mekyska, J., Smekal, Z., Safarova, K., Ondrackova, A., Urbanek, T., Havigerova, J. M., Bednarova, J., & Faundez-Zanuy, M. (2020). Advanced parametrization of graphomotor difficulties in school-aged children. Institute of Electrical and Electronics Engineers Access, 8, 112883–112897. doi:10.1109/ACCESS.2020.3003214
- Galli, M., Vimercati, S. L., Stella, G., Caiazzo, G., Norveti, F., Onnis, F., Rigoldi, C., & Albertini, G. (2011). A new approach for the quantitative evaluation of drawings in children with learning disabilities. Research in Developmental Disabilities, 32(3), 1004–1010. doi:10.1016/j.ridd.2011.01.051
- Graham, S., Harris, K. R., & Fink, B. (2000). Is handwriting causally related to learning to write? treatment of handwriting problems in beginning writers. Journal of Educational Psychology, 92(4), 620. doi:10.1037/0022-0663.92.4.620
- Graham, S., Struck, M., Santoro, J., & Berninger, V. W. (2006). Dimensions of good and poor hand-writing legibility in first and second graders: Motor programs, visual–spatial arrangement, and letter formation parameter setting. Developmental Neuropsychology, 29(1), 43–60. doi:10.1207/ s15326942dn2901_4
- Graham, S., & Weintraub, N. (1996, mar). A review of handwriting research: Progress and prospects from 1980 to 1994. Educational Psychology Review, 8(1), 7–87. doi:10.1007/BF01761831
- Graham, S., Weintraub, N., & Berninger, V. W. (1998). The relationship between handwriting style and speed and legibility. The Journal of Educational Research, 91(5), 290–297. doi:10.1080/ 00220679809597556
- Gravetter, F. J., Wallnau, L. B., Forzano, L.-A. B., & Witnauer, J. E. (2020). Essentials of statistics for the behavioral sciences. Boston, MA: Cengage Learning.Hamstra-Bletz, L., & Blöte, A. W. (1993). A longitudinal study on dysgraphic handwriting in primary school. Journal of Learning Disabilities, 26(10), 689–699. doi:10.1177/002221949302601007
- Hamstra-Bletz, L., DeBie, J., & Den Brinker, B. P. L. M. (1987). Concise evaluation scale for children's handwriting. Lisse: Swets, 1, 623–662.
- Hawke, J. L., Olson, R. K., Willcut, E. G., Wadsworth, S. J., & DeFries, J. C. (2009, aug). Gender ratios for reading difficulties. Dyslexia, 15(3), 239–242. doi:10.1002/dys.389  JASP Team. (2022).
- JASP (Version 0.16.2)[Computer Software]. Retrieved from https://jasp-stats.org/
- Jucovičová, D. (2014). Specifické poruchy učení a chování. Prague, Czech Republic: Univerzita Karlova, Pedagogická fakulta.Kaiser, M.-L., Albaret, J.-M., & Doudin, P.-A. (2009). Relationship between visual-motor integration, eye-hand coordination, and quality of handwriting. Journal of Occupational Therapy, Schools, & Early Intervention, 2(2), 87–95. doi:10.1080/19411240903146228
- Kandel, S., Lassus-Sangosse, D., Grosjacques, G., & Perret, C. (2017, may). The impact of develop-mental dyslexia and dysgraphia on movement production during word writing. Cognitive Neuropsychology, 34(3–4), 219–251. doi:10.1080/02643294.2017.1389706

- Kandel, S., Peereman, R., Grosjacques, G., & Fayol, M. (2011). For a psycholinguistic model of handwriting production: Testing the syllable-bigram controversy. Journal of Experimental Psychology: Human Perception and Performance, 37(4), 1310–1322. doi:10.1037/a0023094
- Kao, H. S., Hoosain, R., & Van Galen, G. P. (1986). Graphonomics: Contemporary research in hand-writing. Amsterdam, Netherlands: Elsevier.Karlsdottir, R., & Stefansson, T. (2002, apr). Problems in developing functional handwriting. Perceptual and Motor Skills, 94(2), 623–662. doi:10.2466/pms.2002.94.2.623
- Katusic, S. K., Colligan, R. C., Weaver, A. L., & Barbaresi, W. J. (2009, may). The forgotten learning disability: Epidemiology of written-language disorder in a population-based birth cohort (1976– 1982), rochester, minnesota. Pediatrics, 123(5), 1306–1313. doi:10.1542/peds.2008-2098
- Khalid, P. I., Yunus, J., & Adnan, R. (2010, jan). Extraction of dynamic features from hand drawn data for the identification of children with handwriting difficulty. Research in Developmental Disabilities, 31(1), 256–262. doi:10.1016/j.ridd.2009.09.009
- Kushki, A., Schwellnus, H., Ilyas, F., & Chau, T. (2011, may). Changes in kinetics and kinematics of handwriting during a prolonged writing task in children with and without dysgraphia. Research in Developmental Disabilities, 32(3), 1058–1064. doi:10.1016/j.ridd.2011.01.026
- McCloskey, M., & Rapp, B. (2017, may). Developmental dysgraphia: An overview and framework for research. Cognitive Neuropsychology, 34(3–4), 65–82. doi:10.1080/02643294.2017.1369016
- Medwell, J., Strand, S., & Wray, D. (2009, sep). The links between handwriting and composing for y6 children. Cambridge Journal of Education, 39(3), 329–344. doi:10.1080/03057640903103728
- Mekyska, J., Faundez-Zanuy, M., Mzourek, Z., Galaz, Z., Smekal, Z., & Rosenblum, S. (2016). Identification and rating of developmental dysgraphia by handwriting analysis. IEEE Transactions on Human-Machine Systems, 47(2), 235–248. doi:10.1109/THMS.2016.2586605
- Mekyska, J., Faundez-Zanuy, M., Mzourek, Z., Galaz, Z., Smekal, Z., & Rosenblum, S. (2017, apr). Identification and rating of developmental dysgraphia by handwriting analysis. IEEE Transactions on Human-Machine Systems, 47(2), 235–248. doi:10.1109/THMS.2016.2586605
- Mekyska, J., Galaz, Z., Safarova, K., Zvoncak, V., Mucha, J., Smekal, Z., Ondrackova, A., Urbanek, T., Havigerova, J. M., Bednarova, J., & Faundez-Zanuy, M. (2019). Computerised assessment of graphomotor difficulties in a cohort of school-aged children. In 2019 11th international congress on ultra-modern telecommunications and control systems and workshops (icumt), Dublin, Ireland (pp. 1–6).
- Meulenbroek, R. G., & Van Galen, G. P. (1986). Movement analysis of repetitive writing behaviour of first, second and third grade primary school children. In H. S. Kao, R. Hoosain, & G. P. Van Galen, (Eds.), Advances in psychology (Vol. 37, pp. 71–92). Amsterdam, Netherlands: Elsevier.
- Mucha, J., Mekyska, J., Zvoncak, V., Galaz, Z., & Smekal, Z. (2022). HandAQUS – handwriting acquisition software. https://github.com/BDALab/HandAQUS. GitHub.
- Nicolson, R. I., & Fawcett, A. J. (2011, jan). Dyslexia, dysgraphia, procedural learning and the cerebellum. Cortex; a Journal Devoted to the Study of the Nervous System and Behavior, 47(1), 117–127. doi:10.1016/j.cortex.2009.08.016

- O'Donnell, E. H., & Colvin, M. K. (2019). Disorders of written expression. In H. K. Wilson, & E. B. Braaten, (Eds.), The massachusetts general hospital guide to learning disabilities (pp. 59–78). Cham, Switzerland: Springer.
- O'Hare, A. E., & Brown, J. K. (1989, March). Childhood dysgraphia. part 1. An illustrated clinical classification. Child: Care, Health and Development, 15(2), 79–104. doi:10.1111/j.1365-2214.1989.
- Overvelde, A., & Hulstijn, W. (2011). Handwriting development in grade 2 and grade 3 primary school children with normal, at risk, or dysgraphic characteristics. Research in Developmental Disabilities, 32(2), 540–548. doi:10.1016/j.ridd.2010.12.027
- Parush, S., Lifshitz, N., Yochman, A., & Weintraub, N. (2010). Relationships between handwriting components and underlying perceptual-motor functions among students during copying and dictation tasks. OTJR: Occupation, Participation, Health, 30(1), 39–48. doi:10.3928/15394492- 20091214-06
- Parush, S., Pindak, V., Hahn-Markowitz, J., & Mazor-Karsenty, T. (1998). Does fatigue influence children's handwriting performance? Work, 11(3), 307–313. doi:10.3233/WOR-1998-11307
- Paz-Villagrán, V., Danna, J., & Velay, J.-L. (2014). Lifts and stops in proficient and dysgraphic handwriting. Human Movement Science, 33, 381–394. doi:10.1016/j.humov.2013.11.005
- Pellizzer, G., & Zesiger, P. (2009, mar). Hypothesis regarding the transformation of the intended direction of movement during the production of graphic trajectories: A study of drawing movements in 8- to 12-year-old children. Cortex; a Journal Devoted to the Study of the Nervous System and Behavior, 45(3), 356–367. doi:10.1016/j.cortex.2008.04.008
- Pokorná, V. (2004). Teorie a náprava vỳvojovỳch poruch učení a chování. Prague, Czech Republic: PORTÁL sro.Prunty, M., & Barnett, A. L. (2017, may). Understanding handwriting difficulties: A comparison of children with and without motor impairment. Cognitive Neuropsychology, 34(3–4), 205–218. doi:10.1080/02643294.2017.1376630
- Razian, M. A., Fairhurst, M. C., & Hoque, S. (2004). Effect of dynamic features on diagnostic testing for dyspraxia. In International conference on computers for handicapped persons (pp. 1039–1046). Paris, France.Rodríguez, C., & Villarroel, R. (2016, mar). Predicting handwriting difficulties through spelling processes. Journal of Learning Disabilities, 50(5), 504–510. doi:10.1177/0022219416633863
- Romani, C., Ward, J., & Olson, A. (1999). Developmental surface dysgraphia: What is the underlying cognitive impairment? The Quarterly Journal of Experimental Psychology Section A, 52(1), 97–128. doi:10.1080/713755804
- Rosenblum, S. (2008). Development, reliability, and validity of the handwriting proficiency screening questionnaire (hpsq). The American Journal of Occupational Therapy, 62(3), 298–307. doi:10.5014/ ajot.62.3.298
- Rosenblum, S. (2018, apr). Inter-relationships between objective handwriting features and exec- utive control among children with developmental dysgraphia. Public Library of Science ONE, 13(4), e0196098. doi:10.1371/journal.pone.0196098
-Rosenblum, S., Aloni, T., & Josman, N. (2010, mar). Relationships between handwriting performance and organizational abilities among children with and without dysgraphia: A preliminary study. Research in Developmental Disabilities, 31(2), 502–509. doi:10.1016/j.ridd.2009.10.016

- Rosenblum, S., Chevion, D., & Weiss, P. L. T. (2006, jan). Using data visualization and signal processing to characterize the handwriting process. Pediatric Rehabilitation, 9(4), 404–417. doi:10.1080/ 13638490600667964
- Rosenblum, S., & Dror, G. (2016). Identifying developmental dysgraphia characteristics utilizing handwriting classification methods. IEEE Transactions on Human-Machine Systems, 47(2), 293–298. doi:10.1109/THMS.2016.2628799
- Rosenblum, S., & Dror, G. (2017, apr). Identifying developmental dysgraphia characteristics utilizing handwriting classification methods. IEEE Transactions on Human-Machine Systems, 47(2), 293–298. doi:10.1109/THMS.2016.2628799
- Rosenblum, S., Dvorkin, A. Y., & Weiss, P. L. (2006, oct). Automatic segmentation as a tool for examining the handwriting process of children with dysgraphic and proficient handwriting. Human Movement Science, 25(4–5), 608–621. doi:10.1016/j.humov.2006.07.005
-Rosenblum, S., & Gafni-Lachter, L. (2015). Handwriting proficiency screening questionnaire for children (hpsq–c): Development, reliability, and validity. The American Journal of Occupational Therapy, 69(3), p69032200301–p69032200309. doi:10.5014/ajot.2015.014761
-Rosenblum, S., Goldstand, S., & Parush, S. (2006, January). Relationships among biomechanical ergonomic factors, handwriting product quality, handwriting efficiency, and computerized handwriting process measures in children with and without handwriting difficulties. The American Journal of Occupational Therapy, 60(1), 28–39. doi:10.5014/ajot.60.1.28
- Rosenblum, S., Parush, S., & Weiss, P. L. (2003, March). Computerized temporal handwriting characteristics of proficient and non-proficient handwriters. The American Journal of Occupational Therapy, 57(2), 129–138. doi:10.5014/ajot.57.2.129
- Rosenblum, S., & Roman, H. E. (2009, mar). Fluctuation analysis of proficient and dysgraphic handwriting in children. EPL (Europhysics Letters), 85(5), 58007. doi:10.1209/0295-5075/85/58007
- Rosenblum, S., Weiss, P. L., & Parush, S. (2003). Product and process evaluation of handwriting difficulties. Educational Psychology Review, 15(1), 41–81. doi:10.1023/A:1021371425220 Rosenblum, S., Weiss, P. L., & Parush, S. (2004). Handwriting evaluation for developmental dysgraphia: Process versus product. Reading and Writing, 17(5), 433–458. doi:10.1023/B:READ. 0000044596.91833.55
- Roston, K. L., Hinojosa, J., & Kaplan, H. (2008a). Using the minnesota handwriting assessment and handwriting checklist in screening first and second graders' handwriting legibility. Journal of Occupational Therapy, Schools, & Early Intervention, 1(2), 100–115. doi:10.1080/ 19411240802312947
- Roston, K. L., Hinojosa, J., & Kaplan, H. (2008b, sep). Using the minnesota handwriting assessment and handwriting checklist in screening first and second graders' handwriting legibility. Journal of Occupational Therapy, Schools, & Early Intervention, 1(2), 100–115. doi:10.1080/ 19411240802312947
- Rubin, N., & Henderson, S. E. (1982). Two sides of the same coin: Variations in teaching methods and failure to learn to write. Special Education: Forward Trends, 9(4), 17–24. doi:10.1111/j.1467-8578. 1982.tb00576.x
- Šafárová, K., Mekyska, J., & Zvončák, V. (2021). Developmental dysgraphia: A new approach to diagnosis. The International Journal of Assessment and Evaluation, 28(1), 143–160. doi:10.18848/ 2327-7920/CGP/v28i01/143-160

- Šafárová, K., Mekyska, J., Zvončák, V., Galáž, Z., Francová, P., Čechová, B., Losenická, B., Smékal, Z., Urbánek, T., Havigerová, J. M., & Rosenblum, S. (2020, jan). Psychometric properties of screening questionnaires for children with handwriting issues. Frontiers in Psychology 10. 10.3389/fpsyg. 2019.02937 .
- San Luciano, M., Wang, C., Ortega, R. A., Yu, Q., Boschung, S., Soto-Valencia, J., Bressman, S. B., Lipton, R. B., Pullman, S., & Saunders- Pullman, R. (2016). Digitized spiral drawing: A possible biomarker for early Parkinson's disease. Public Library of Science ONE, 11(10), e0162799. doi:10.1371/journal. pone.0162799
- Schoemaker, M., Schellekens, J., Kalverboer, A., & Kooistra, L. (1994). Pattern drawing by clumsy children: A problem of movement control. Contemporary Issues in the Forensic, Developmental, and Neurological Aspects of Handwriting, 1, 45–64. Schoemaker, M., & Smits-Engelsman, B. (1997). Dysgraphic children with and without a generalized motor problem: Evidence for subtypes. In Igs 1997 proceedings: Eighth biennial conference, Genoa, Italy (pp. 11–12).
- Schwellnus, H., Carnahan, H., Kushki, A., Polatajko, H., Missiuna, C., & Chau, T. (2012, nov). Effect of pencil grasp on the speed and legibility of handwriting in children. The American Journal of Occupational Therapy, 66(6), 718–726. doi:10.5014/ajot.2012.004515
- Simner, M. L., & Eidlitz, M. R. (2000). Work in progress: Towards an empirical definition of developmental dysgraphia: Preliminary findings. Canadian Journal of School Psychology, 16(1), 103–110. doi:10.1177/082957350001600108 Smith-Zuzovsky, N., & Exner, C. E. (2004, jul). The effect of seated positioning quality on typical 6- and 7-year-old children's object manipulation skills. The American Journal of Occupational Therapy, 58(4), 380–388. doi:10.5014/ajot.58.4.380
- Smits-Engelsman, B. C., Niemeijer, A. S., & van Galen, G. P. (2001). Fine motor deficiencies in children diagnosed as dcd based on poor graphomotor ability. Human Movement Science, 20(1–2), 161–182. doi:10.1016/S0167-9457(01)00033-1
- Smits-Engelsman, B. C., & Van Galen, G. P. (1997). Dysgraphia in children: Lasting psychomotor deficiency or transient developmental delay? Journal of Experimental Child Psychology, 67(2), 164–184. doi:10.1006/jecp.1997.2400
- Snowling, M. J. (2005, sep). Specific learning difficulties. Psychiatry, 4(9), 110–113. doi:10.1383/psyt. 2005.4.9.110 Søvik, N., Arntzen, O., & Thygesen, R. (1987). Relation of spelling and writing in learning disabilities. Perceptual and Motor Skills, 64(1), 219–236. doi:10.2466/pms.1987.64.1.219
- Søvik, N., Flem Mæland, A., & Karlsdottir, R. (1989). Contextual factors and writing performance of 'normal'and dysgraphic children. In C. Y. Suen, R. Plamondon, & M. L. Simner (Eds.), Computer recognition and human production of handwriting (pp. 333–347). Singapore: World Scientific. Trochim, W. M., & Donnelly, J. P. (2001). Research methods knowledge base (Vol. 2). New York: Atomic Dog Pub. Macmillan Publishing Company. Tseng, M. H., & Chow, S. M. K. (2000, jan). Perceptual-motor function of school-age children with slow handwriting speed. The American Journal of Occupational Therapy, 54(1), 83–88. doi:10.5014/ ajot.54.1.83
- Tucha, O., Tucha, L., & Lange, K. W. (2008). Graphonomics, automaticity and handwriting assessment. Literacy, 42(3), 145–155. doi:10.1111/j.1741-4369.2008.00494.x
- Vaivre-Douret, L., Lopez, C., Dutruel, A., & Vaivre, S. (2021). Phenotyping features in the genesis of pre-scriptural gestures in children to assess handwriting developmental levels. Scientific Reports, 11(1), 1–13. doi:10.1038/s41598-020-79315-w

33- van Galen, G. P. (1991, may). Handwriting: Issues for a psychomotor theory. Human Movement Science, 10(2–3), 165–191. doi:10.1016/0167-9457(91)90003-G
- Van Galen, G. P., Portier, S. J., Smits-Engelsman, B. C., & Schomaker, L. R. (1993). Neuromotor noise and poor handwriting in children. Acta Psychologica, 82(1–3), 161–178. doi:10.1016/0001- 6918(93)90010-O
- Van Gemmert, A. W., & Contreras-Vidal, J. L. (2015). Graphonomics and its contribution to the field of motor behavior: A position statement. Human Movement Science, 43, 165–168. Van Gemmert, A. W., & Teulings, H.-L. (2006). Advances in graphonomics: Studies on fine motor control, its development and disorders. Human Movement Science, 25(4–5), 447–453. doi:10.1016/ j.humov.2006.07.002
- Van Waelvelde, H., Hellinckx, T., Peersman, W., & Smits-Engelsman, B. C. (2012). Sos: A screening instrument to identify children with handwriting impairments. Physical & Occupational Therapy in Pediatrics, 32(3), 306–319.
- Wann, J., & Jones, J. (1986). Space-time invariance in handwriting: Contrasts between primary school children displaying advanced or retarded handwriting acquisition. Human Movement Science, 5 (3), 275–296. doi:10.1016/0167-9457(86)90032-1
- Wann, J., Wing, A. M., & Sõvik, N. (Eds.). (1991). Development of graphic skills: Research perspectives and educational implications. London, UK: Academic Press. Weintraub, N., & Graham, S. (1998). Writing legibly and quickly: A study of children's ability to adjust their handwriting to meet common classroom demands. Learning Disabilities Research & Practice, 13(3), 146–152.
- Wicki, W., Lichtsteiner, S. H., Geiger, A. S., & Müller, M. (2014, jan). Handwriting fluency in children. Swiss Journal of Psychology, 73(2), 87–96. doi:10.1024/1421-0185/a000127
- World Health Organization. (2016). International Statistical Classification of Diseases and Related Health Problems (10th ed.). https://icd.who.int/browse10/2016/en
- Zelinková, O. (2015). Poruchy učení: dyslexie, dysgrafie, dysortografie, dyskalkulie, dyspraxie, ADHD. Prague, Czech Republic: Portál.
- Ziviani, J. (1984). Some elaborations on handwriting speed in 7-to 14-year-olds. Perceptual and Motor Skills, 58(2), 535–539. doi:10.2466/pms.1984.58.2.535
- Ziviani, J., & Wallen, M. (2006). Chapter 11 – the development of graphomotor skills. In A. Henderson & C. Pehoski (Eds.), Hand function in the child (second edition) (Second Edition ed., pp. 217–236). Saint Louis: Mosby.
**Appendices**

List of appendices:
 A1 reviewed scales
A2 reviewed web pages.
A3 the graphomotor part of the protocol
A4 the template consisting of lines printed 20 mm apart.
A5 the template consisting of lines printed 15 mm apart.
A6 the protocol used during simulation of manifestations.
A7 tables with normative data